\pdfoutput=1

\documentclass[11pt]{article}

\usepackage[table]{xcolor} 
\usepackage[final]{acl}

\usepackage{times}
\usepackage{latexsym}

\usepackage[T1]{fontenc}

\usepackage[utf8]{inputenc}

\usepackage{microtype}

\usepackage{inconsolata}

\usepackage{graphicx}
\usepackage{booktabs}

\usepackage{amsfonts}

\title{The Automated Verification of Textual Claims (AVeriTeC) Shared Task}

\author{%
Michael Schlichtkrull\textsuperscript{1,2}, Yulong Chen\textsuperscript{2}, Chenxi Whitehouse\textsuperscript{2,5}, Zhenyun Deng\textsuperscript{2},\\ \textbf{Mubashara Akhtar\textsuperscript{4}, Rami Aly\textsuperscript{2}, Zhijiang Guo\textsuperscript{2}, Christos Christodoulopoulos\textsuperscript{3},}\\ \textbf{Oana Cocarascu\textsuperscript{4}, Arpit Mittal\textsuperscript{5}, James Thorne\textsuperscript{6}, Andreas Vlachos\textsuperscript{2}} \\
\textsuperscript{1}Queen Mary University of London,
\textsuperscript{2}University of Cambridge, \\
\textsuperscript{3}Amazon AGI,
\textsuperscript{4}King's College London, 
\textsuperscript{5}Meta,
\textsuperscript{6}KAIST\\
  \texttt{m.schlichtkrull@qmul.ac.uk, \{yc632,cj507,zd302,rmya2,zg283,av308\}@cam.ac.uk}\\
   \texttt{chrchrs@amazon.co.uk, \{mubashara.akhtar,oana.cocarascu\}@kcl.ac.uk} \\
   \texttt{thorne@kaist.ac.kr, arpitmittal@meta.com} \\
}

\begin{document}
\maketitle
\begin{abstract}
The Automated Verification of Textual Claims (\textsc{AVeriTeC}) shared task asks participants to retrieve evidence and predict veracity for real-world claims checked by fact-checkers. Evidence can be found either via a search engine, or via a knowledge store provided by the organisers. Submissions are evaluated using the \textsc{AVeriTeC} score, which considers a claim to be accurately verified if and only if both the verdict is correct and retrieved evidence is considered to meet a certain quality threshold.
The shared task received 21 submissions, 18 of which surpassed our baseline. The winning team was TUDA\_MAI with an \textsc{AVeriTeC} score of 63\%. In this paper we describe the shared task, present the full results, and highlight key takeaways from the shared task.
\end{abstract}

\section{Introduction}

Automated fact-checking (AFC) has been proposed as an assistive tool for beleaguered fact-checkers~\citep{cohen2011computational, vlachos-riedel-2014-fact}, whose work is crucial for limiting misinformation~\citep{lewandowsky_debunking_2020}. This has inspired applications in journalism~\citep{miranda2019newsroom, fullfact2020, nakov2021assisting} and other domains, e.g.\ science~\citep{wadden-etal-2020-fact}. Substantial progress has been made on common benchmarks, such as FEVER~\citep{thorne-etal-2018-fever} and MultiFC \citep{augenstein-etal-2019-multifc}. Nevertheless, existing resources have recently come under criticism. Many datasets (for example,~\citet{thorne-etal-2018-fever, schuster-etal-2021-get, aly-etal-2021-fact}) contain purpose-made claims derived e.g.\ from Wikipedia, and are thus not representative of real-world use cases. Datasets that \textit{do} contain real-world claims either lack evidence annotation~\citep{wang-2017-liar}, or suffer issues resulting from superficial automated evidence annotation~\citep{glockner-etal-2022-missing}.

\begin{figure}[t]
    \centering
    \includegraphics[scale=0.7]{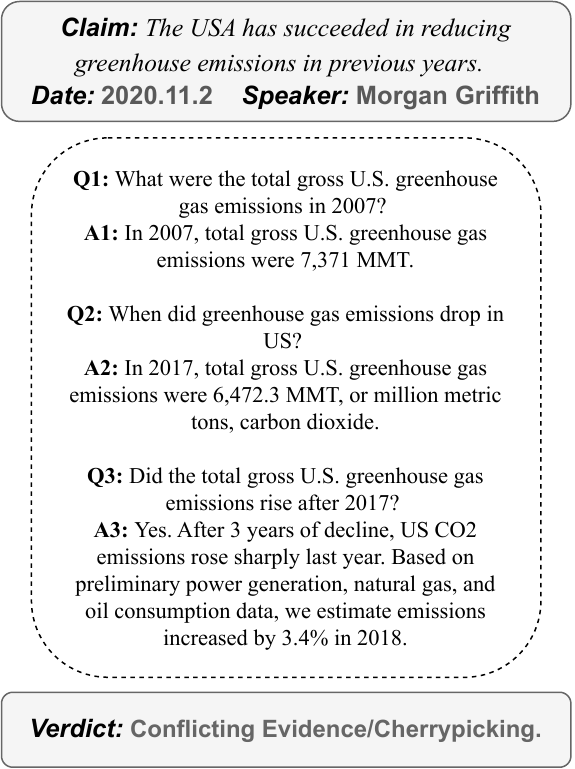}
    \caption{Example instance from \textsc{AVeriTeC}. Given a claim and associated metadata, participating systems must first retrieve appropriate evidence. Then, they must output a verdict for the claim given that evidence.}
    \label{fig:dataset example}
\end{figure}


The \textsc{AVeriTeC} dataset was constructed to overcome these limitations
~\citep{schlichtkrull2023averitec}. \textsc{AVeriTeC} combines real-world claims with evidence from the web. The 
process of evidence retrieval is broken down into question generation and answering, providing a structured representation of the evidential reasoning process. The annotation process for \textsc{AVeriTeC} was designed to ensure (1) that claims are understandable independently of the fact-checking articles they were sourced from, (2) that the evidence given is sufficient to support the verdicts, and (3) that all evidence used would have been available on the web before the claim was made. This avoids common problems found in previous datasets~\citep{ousidhoum-etal-2022-varifocal, glockner-etal-2022-missing}.

\textsc{AVeriTeC} consists originally of 4,568 examples, collected from 50 fact-checking organizations using the Google FactCheck Claim Search API\footnote{\url{https://toolbox.google.com/factcheck/apis}, available under a \textsc{CC-BY-4.0} license.}; itself based on ClaimReview\footnote{\url{https://www.claimreviewproject.com/}}. To ensure that systems are evaluated on unseen data, we expanded the (hidden) test set with a further 1,215 claims for the shared task, bringing the total dataset size to 5,783. We furthermore released a ``knowledge store'' containing, for each claim in the training, development, and test splits, documents which can be used as evidence for that claim. This was done to prevent participants from being limited by the prohibitive cost of the search API we used for evidence retrieval in the original paper~\citep{schlichtkrull2023averitec}. We also developed an updated version of the baseline for the shared task, which uses the knowledge store. Participants in the shared task were allowed to use evidence from the knowledge store, use a search engine on their own, or combine the two options. Our dataset and baseline are available under a \textsc{CC-BY-NC-4.0} license at \url{https://fever.ai/dataset/averitec.html}.

This paper presents a description of the task and dataset, the final test phase leaderboard. We also summarise the submitted system description papers, drawing out commonalities, differences, and lessons. We furthermore carry out additional analysis of the shared task results, including human evaluation. Finally, we reflect on the task, deriving lessons for future work -- and further shared tasks -- on automated fact-checking. The shared task received 21 submissions. The winning team, TUDA\textunderscore MAI, achieved a score of 63\%, a very significant improvement on the 11\% achieved by the baseline system. Nevertheless, there are still plenty of opportunities for further improvement. During the process, we identified an issue with the evidence set provided for participants, which for some claims in the second half of the dataset contained fact-checking articles written by humans about those claims. We release an updated knowledge store at \url{https://fever.ai/dataset/averitec.html}, where these articles have been removed. We leave open an evaluation page corresponding to the \textit{new} knowledge store\footnote{Also available at \url{https://fever.ai/dataset/averitec.html}} so that future work can build upon the advances made in this shared task.

\section{Task Description}

Participants are given claims and associated metadata, such as the publication date (see Figure~\ref{fig:dataset example}). Based on this, they must retrieve \textit{evidence} for or against the claims. In the gold annotation, this evidence is broken down into question-answer pairs, naturally enabling multi-hop reasoning. We do not restrict participants to providing evidence in this format, although given the METEOR-based evaluation setup most participants found it beneficial to follow it. When submitting test set predictions, we also required participants to include a URL to an external website for each piece of evidence, corresponding to a webpage providing \textit{backing}. Finally, based on the evidence, participants must predict whether a veracity label from the set \textit{supported, refuted, not enough evidence}, or \textit{conflicting evidence/cherrypicking}. Unlike the original \textsc{AVeriTeC} dataset, we did not require participants to submit a justification for the verdict.

\subsection{Dataset}
Participants are asked to use the public \textsc{AVeriTeC} data for training and validating their systems. 
To ensure a fairer and more robust evaluation, we constructed a new test set consisting of 1,215 claims, which temporally succeed the original claims, in addition to the original 1000 hidden test set claims of \textsc{AVeriTeC}. Like the original test set, these will remain hidden so as to enable future work on the dataset.

\begin{table*}[ht]
    \centering
    \scalebox{0.75}{
\addtolength{\tabcolsep}{0pt}
    \begin{tabular}{l|rrrr}
    \toprule
    \textbf{Split} &  \textbf{Train} &  \textbf{Dev} &  \textbf{Test (old)} & \textbf{Test (new)}\\
    \midrule
    Claims &  3,068 & 500 & 1,000 & 1,215\\
    Question / Claim & 2.60 & 2.57 & 2.57 & 2.89 \\
    Re-annotated (\%) & 28.1 & 24.4 & 25.1 &20.0 \\
    End date & 25-08-2020 & 31-10-2020& 22-12-2021 & 13-08-2023 \\
    Labels (S/R/C/N) & 27.6/56.8/6.4/9.2 & 24.4/61.0/7.6/7.0 & 25.5/62.0/6.3/6.2 & 17.3/66.5/4.1/12.1\\
    Types (PS/NC/EPC/QV/CC) & 7.8/33.7/57.8/9.6/11.5 & 5.8/23.8/61.4/13.8/10.8 & 7.0/21.9/69.8/7.7/11.9 & 3.5/24.3/71.9/5.2/16.1 \\
    Strategies (WE/NCP/FR/EC/SS) & 78.8/30.6/6.6/29.9/3.6 & 88.6/19.0/7.4/27.4/2.0 & 88.0/19.2/7.7/29.6/1.8 & 82.4/22.6/10.0/37.6/4.0\\
    \bottomrule
    \end{tabular}
    }
    \caption{Statistics for the new test set. For better comparison, we present the statistics for the original dataset. The Labels (\%)  are Supported (S), Refuted (R), Conflicting Evidence/Cherry-picking (C), and Not Enough Evidence (N). The Claim Types (\%) are Position Statement (PS), Numerical Claim (NC), Event/Property Claim (EPC), Quote Verification (QV), and Causal Claim (CC). The Fact-checker strategies (\%) are Written Evidence (WE),
Numerical Comparison (NCP), Fact-checker Reference (FR), Expert Consultation (EC) and Satirical Source (SS). Note that we for simplicity omitted very low-frequent fact-checker strategies, e.g., Geo-location (0.3\%).}
    \label{tab:stat_new_test}
\end{table*}

\paragraph{Annotation of New Test Set}
We first collect 2,000 real-world fact-checking articles online from ClaimReview,
same source as \textsc{AVeriTeC}.
Then, we follow the same 5-phase annotation guideline of \citet{schlichtkrull2023averitec}.

First, given a fact-checking article, an annotator identifies its main claim, collects metadata about it and normalizes the claim by enriching it with necessary context, making it context-independent.
Second, given the normalized claim, another annotator generates questions and answers (QAs) with the help of the fact-checking article and the web, and gives a verdict label for the claim.
Third, given only the QAs as evidence, a different annotator selects a verdict label for the claim and provides a justification for their choice.
At this point, we compare the verdict labels annotated by different annotators.
If the labels match, we consider the evidence is sufficient for predicting the veracity; otherwise, we repeat the last two phases as our fourth and fifth phases, respectively.
If the labels given by the fourth and fifth annotators still do not match, we discard this instance.
In this way, we obtain 1,215 new instances. Each is annotated with a normalized claim, meta-data, QA pairs as evidence, a verdict label and a justification for it.
For the detailed annotation guidelines and procedures, please refer to \citet{schlichtkrull2023averitec}.

To ensure high quality, we train our annotators before formal annotation.
For each phase, annotators are first asked to annotate 10 instances.
We then provide feedback and highlight their most frequent and common mistakes.
They are then asked to annotate another 10 instances. 
We select qualified annotators based on their performance on 3 tasks: (1) claim type and fact-checking strategies over $70\%+$ $F$-1 scores; (2)
$2+$ QA pairs per claim; (3) veracity prediction over $50\%+$ accuracy.
These criteria are based on empirical consideration from the earlier \textsc{AVeriTeC} annotation~\citep{schlichtkrull2023averitec}.
Finally, we selected 12 qualified annotators from 34 participants.

\paragraph{Comparison between Original and New Test Sets}

\autoref{tab:stat_new_test} presents the statistics of our new test set in comparison with the original \textsc{AVeriTeC} dataset. 
Our new test set (with claims up to 2023) is temporally further removed from the training set (ending in 2020).
As such, there can be a domain shift between new and old data, regarding the fact-checking content.
However, 
the majority (66.5\%) of claim labels are \emph{refuted}, which is consistent with previous data.
Additionally, the distributions of claim labels, claim types and fact-checking strategies are largely similar in terms of their proportions.
The new test set has slightly more questions per claim compared to the original one, indicating that the annotation process was at least as thorough. 

\subsection{Knowledge Store}
As mentioned in~\citet{schlichtkrull2023averitec}, reliance on the Google search API made the original baseline prohibitively expensive. Thus, to mitigate the cost, we released a \textit{knowledge store} along with the shared task. The knowledge store contains a collection of potentially useful evidence documents for each claim, obtained via Google search.

We collected the knowledge store using a process inspired by our original baseline. We extracted a variety of search queries using ChatGPT\footnote{We used \texttt{gpt-3.5-turbo-0125}.}, based on the claim, gold questions, and gold answers. We further used \textit{distractor queries} created by changing entities, dates, and events in the claim, in order to add plausible -- but irrelevant -- documents to the knowledge store. All queries can be seen in Appendix~\ref{appendix:search_query_prompts}. For each query, we collected every URL returned on the first page of the Google Search API. We used the same temporal restrictions as in \citet{schlichtkrull2023averitec}, ensuring that the included documents would have been available on the web before the claim was made. We also included the annotator-selected evidence documents selected for each claim. We deduplicated and shuffled the documents corresponding to each claim.


We provided the URL for each document, as well as a text version scraped using \texttt{trafilatura} ~\citep{barbaresi-2021-trafilatura}. The knowledge store includes text scraped from PDF URLs, a step omitted in \citet{schlichtkrull2023averitec}. Furthermore, for the train and development splits (but not test), we made available the specific Google search query used for each document, as well as the category (see Table~\ref{table:knowledge_store_prompts}). The average claim has 955 associated documents, each of which have on average of 6,095 tokens. The most common URL domains for knowledge store documents are, in order, the National Center for Biotechnology Information (NCBI), Wikipedia, Quora, the New York Times, and CNN.

The knowledge store allowed participants to compete without access to a paid search engine. Further, it allowed inexpensive experimentation with a variety of different retrieval strategies. Our construction process for the knowledge store relies on information not available normally to participants, such as the gold question-answer pairs. We found that these were necessary to ensure that good, relevant evidence was included. At the same time, relying on the knowledge store complicates the finding of alternative evidence paths to the one used by our annotators. Exploring alternative evidence paths was easier for systems which directly integrated their own search engine. As such, there were upsides to both strategies. 

\subsection{Baseline}

Our baseline closely follows the approach described in \citet{schlichtkrull2023averitec}, with the main difference being that, instead of requiring direct access to the paid Google Search API, we use the aforementioned knowledge store. This adjustment aims to reduce the costs of participating in the Shared Task. 

Our baseline consists of the following steps. (1) We parse the scraped text into sentences and rank their similarity to the claim using BM25~\citep{robertson2009bm25}, retaining the top 100 sentences per claim. (2) Questions-answer (QA) pairs are generated for these top 100 sentences using BLOOM,\footnote{We used \texttt{bigscience/bloom-7b1}.} with the 10 most similar claim-QA pairs from the training set used as in-context examples. (3) The QA pairs are then re-ranked using a pretrained BERT model as described in \citet{schlichtkrull2023averitec}. (4) Finally, using the top-3 QA pairs as evidence, we predict the veracity label of the claim with another pretrained BERT model, as detailed in \citet{schlichtkrull2023averitec}.

The baseline results are shown in \autoref{tab:overall_results}. We note that on both the development set, the old test set, and the new test set, the shared task baseline and the baseline from \citet{schlichtkrull2023averitec} perform similarly. Further details regarding the implementation, knowledge store, and pretrained BERT models are available at {\url{https://huggingface.co/chenxwh/AVeriTeC}}.

\subsection{Evaluation}
\label{ssec:evaluation}

The primary evaluation metric for the shared task is \textsc{AVeriTeC} score, discussed in depth in \citet{schlichtkrull2023averitec}. We first compute results for question generation and question-answer generation using Hungarian METEOR score. That is, we use the Hungarian Algorithm~\citep{kuhn1955hungarian} to find an optimal matching of generated text to reference text in terms of METEOR score. Formally, let $X : \hat{Y} \times Y \to \{0,1\}$ be a boolean function denoting the assignment between the first 10 generated question-answer pairs (or questions only) $\hat{Y}$ and the reference question-answer pairs (or questions only) $Y$. Then, the Q + A score (or Q only score) $u$ is calculated as:
\begin{equation}
    u_f(\hat{Y}, Y) = \frac{1}{|Y|} \max \sum\limits_{\hat{y} \in \hat{Y}} \sum\limits_{y \in Y} f(\hat{y}, y) X(\hat{y}, y)
\end{equation}
where the pairwise scoring function $f: S \times S \to \mathbb{R} $ is METEOR score~\citep{banerjee-lavie-2005-meteor}  using the NLTK implementation~\citep{bird2009natural}.

To compute the \textsc{AVeriTeC} score, we applied a cutoff of $u_f(\hat{Y}, Y) \geq 0.25$ to determine whether adequate evidence has been retrieved, using the Q + A Hungarian METEOR score. Any claim for which this score is lower then $0.25$ receives an \textsc{AVeriTeC} score of $0$. For claims where the evidence score is higher than $0.25$, the \textsc{AVeriTeC} score is defined as the accuracy of the predicted verdict (veracity). As also discussed in \citet{schlichtkrull2023averitec}, both for Q only, Q+A, and \textsc{AVeriTeC} score, if a system provided more than 10 QA pairs, all pairs after the 10th were discarded. We note that QA pairs beyond the 10th can still be input to veracity prediction components, and may as such still be useful to some systems.

\begin{table*}[ht]
    \centering
\begin{tabular}{ll|rrr}
\toprule
Rank & Team Name                     & Q only & Q + A & \textsc{AVeriTeC} @ .25 \\ \midrule
1    & TUDA\_MAI ~\citep{tudamai}      & 0.45   & 0.34  & \textbf{0.63}           \\
2    & HUMANE ~\citep{humane}                 & 0.48   & \textbf{0.35}  & 0.57           \\
3    & CTU AIC ~\citep{ctu-aic}                      & 0.46   & 0.32  & 0.50           \\
4    & Dunamu-ml  ~\citep{dunamu-ml}                   & \textbf{0.49}   & \textbf{0.35}  & 0.50           \\
5    & Papelo ~\citep{papelo}                       & 0.44   & 0.30  & 0.48           \\
6    & UHH  ~\citep{uhh}                         & 0.48   & 0.32  & 0.45           \\
7    & SynApSe ~\citep{synapse}                      & 0.41   & 0.30  & 0.42           \\
8    & arioriAveri ~\citep{arioriaveri} & 0.38   & 0.29  & 0.39           \\
9    & Data-Wizards  ~\citep{data-wizards}                & 0.35   & 0.27  & 0.33           \\
10   & MA-Bros-H ~\citep{mabros}                    & 0.38   & 0.24  & 0.27           \\
11   & mitchelldehaven               & 0.27   & 0.23  & 0.25           \\
12   & SK\_DU  ~\citep{skdu}                      & 0.40   & 0.26  & 0.22           \\
13   & UPS  ~\citep{ups}                         & 0.31   & 0.27  & 0.21           \\
14   & FZI-WIM    ~\citep{fzi-wim}                   & 0.32   & 0.21  & 0.20           \\
15   & KnowComp   ~\citep{knowcomp}                   & 0.32   & 0.21  & 0.18           \\
16   & IKR3-UNIMIB ~\citep{ikr3}            & 0.32   & 0.24  & 0.18           \\
17   & ngetach                       & 0.37   & 0.21  & 0.14           \\
18   & VGyasi                        & 0.38   & 0.22  & 0.12           \\
19   & \textit{Baseline}  & \textit{0.24}   & \textit{0.20}  & \textit{0.11}           \\
20   & InfinityScalers!              & 0.26   & 0.19  & 0.08           \\
21   & AYM                           & 0.13   & 0.12  & 0.06           \\
22   & Factors                       & 0.20   & 0.14  & 0.05           \\
\bottomrule 
\end{tabular}
    \caption{Overall results for the \textsc{AVeriTeC} shared task. Performance is evaluated on the total of 2214 hidden test set examples. Scores are given in Hungarian METEOR for question-only and question-answer performance, and in \textsc{AVeriTeC}-score at evidence cutoff $0.25$ for total performance (see \citet{schlichtkrull2023averitec}).}
    \label{tab:overall_results}
\end{table*}

\section{Results}
\label{sec:results}

\begin{table*}[htb]
\scalebox{0.88}{
\addtolength{\tabcolsep}{0pt}
\rowcolors{2}{white}{gray!15} 
\begin{tabular}{l|lllll}
\toprule
\textbf{Team Name} & \textbf{Evidence} & \textbf{QG} & \textbf{Retrieval} & \textbf{QA} & \textbf{Veracity} \\
\midrule
TUDA\_MAI & KS & GPT-4o & gte\_base\_en\_v1.5 & GPT-4o & GPT-4o \\
HUMANE & KS & Llama-3-8b & \begin{tabular}[c]{@{}l@{}}BM25\\ SFR-embedding-2\\ Llama-3.1-70b\end{tabular} & - & Llama-3.1-70b \\
CTU AIC & KS & GPT-4o & mxbai-large-v1 & GPT-4o & GPT-4o \\
Dunamu-ML & KS & GPT-4 & BM25 & GPT-4 & GPT-4 \\
Papelo & Google & \begin{tabular}[c]{@{}l@{}}T5-large\\ GPT-4o\end{tabular} & - & GPT-4o & GPT-4o \\
UHH & KS & GPT-4o-mini & \begin{tabular}[c]{@{}l@{}}BM25\\ gte\_base\_en\_v1.5\end{tabular} & GPT-4o-mini & Mixtral-8x7B \\
SynApSe & Google & GPT-4o & all-MiniLM-L6-v2 & GPT-4o & \begin{tabular}[c]{@{}l@{}}GPT-4o\\ GPT-3.5\\ Mistral-7B\end{tabular} \\
aioriAveri & KS & GPT-4o & stella\_en\_400M\_v5 & GPT-4o & GPT-4o \\
Data-Wizards & KS & Phi-3-medium & stella\_en\_1.5B\_v5 & Mixtral-8x22B & Mixtral-8x22B \\
MA-Bros-H & KS & Llama-3-70B & BM25 & Llama-3-70B & Llama-3-70B \\
SK\_DU & KS & GPT-4o & \begin{tabular}[c]{@{}l@{}}BM25\\ ms-marco-MiniLM-L-12-v2\end{tabular} & - & deberta-v3-base \\
UPS & KS & T5-large & \begin{tabular}[c]{@{}l@{}}BM25\\ BERT\end{tabular} & - & BERT \\
FZI-WIM & KS & Llama-3-70B & ms-marco-MiniLM-L-12-v2 & \begin{tabular}[c]{@{}l@{}}Llama-3-70B\\ bart-large-mnli\end{tabular} & Llama-3-70B \\
KnowComp & Google & Llama-3-8b & - & Llama-3-8b & Llama-3-8b \\
IKR3-UNIMIB & KS & - & \begin{tabular}[c]{@{}l@{}}BM25\\ ColBERT\end{tabular} & GPT-3.5 & BERT \\
\bottomrule
\end{tabular}
}
\caption{Components used by systems that submitted description papers. Systems are ordered based on AVeriTeC-score (see Table~\ref{tab:overall_results}). - indicates, respectively, that a system directly used claims and nothing else for search queries, that retrieval was done only through a search API with no reranking, and that the answer used was the entire retrieved passage.}
\label{table:components}
\end{table*}

The overall results for the shared task can be seen in Table~\ref{tab:overall_results}. Each of the 21 participating teams were invited to submit a paper to be reviewed in the FEVER workshop -- detailed descriptions for each system can be found in the corresponding papers. 15 system description papers were submitted to the workshop (with a 16th submitted and withdrawn). We analyse the model components discussed in each paper -- see Table~\ref{table:components}. Below, we present our general observations on the techniques used by participants, as reported in their respective system description papers.

\pagebreak

\paragraph{Knowledge Source}

Papelo, SynApSe, and KnowComp relied on the Google Search API as knowledge source, while the remaining systems all used our knowledge store. Participants identified shortcomings in both approaches: the knowledge store is guaranteed to include the gold evidence and can be searched with highly performant embedding methods, whereas the search API  
allows for more freedom in what information can be retrieved (i.e., if generating questions for a different evidence path than the one our annotators used, the knowledge store may not be able to answer those questions). As evidenced by the strong results of Team Papelo, despite the predominance of systems relying on the knowledge store, the Google Search API (with which the knowledge store itself was built) remained a competitive option (see Table~\ref{tab:overall_results}).

One issue identified by several participants was the scraper we used for the knowledge store, based on Trafilatura~\citep{barbaresi-2021-trafilatura}. Papelo identified how, in 297 out of 500 development examples, at least one gold document was not correctly scraped. Dunamu-ML similarly discussed how the scraper did not correctly handle evidence from PDFs and videos. In their submission, Dunamu-ML extended the scraper to extract text and transcripts from PDFs and YouTube videos, and noted that this helped performance. When constructing \textsc{AVeriTeC}, our annotators filtered out claims requiring multimodal reasoning; all claims in the dataset are textual and can be verified through exclusively textual evidence. Nevertheless, the helpfulness of video transcripts suggests that multimodal evidence can be useful even for that scenario.

\begin{table*}[ht]
\centering
\scalebox{0.88}{
\begin{tabular}{l|rrrrrrrr|rrr}
\toprule
Team Name                                     &  Text & PDF & Table & Metadata & Audio & Video & Image & Other & 1 doc & 2 docs & 3+ docs \\ \midrule
TUDA\_MAI                   & \textbf{0.34}                           & 0.35                      & 0.36                            & 0.31                           & 0.31                        & \textbf{0.33}                        & 0.32                                & 0.33                        & 0.39                               & 0.35                               & 0.31                                \\
HUMANE                              & \textbf{0.34}                           & \textbf{0.36}                      & 0.38                            & \textbf{0.32}                           & \textbf{0.34}                        & 0.32                        & 0.33                                & \textbf{0.38}                        & \textbf{0.41}                               & 0.35                               & 0.31                                \\
CTU AIC                              & 0.31                           & 0.33                      & 0.36                            & 0.30                            & 0.26                        & 0.30                         & 0.32                                & 0.35                        & 0.33                               & 0.33                               & 0.29                                \\
Dunamu-ml                                  & \textbf{0.34}                           & \textbf{0.36}                      & \textbf{0.39}                            & 0.31                           & 0.24                        & \textbf{0.33}                        & \textbf{0.34}                                & 0.37                        & 0.40                                & \textbf{0.36}                               & \textbf{0.32}                                \\
Papelo                                     & 0.3                            & 0.31                      & 0.32                            & 0.27                           & 0.22                        & 0.29                        & 0.29                                & 0.3                         & 0.35                               & 0.3                                & 0.27                                \\
UHH                                        & 0.31                           & 0.34                      & 0.36                            & 0.29                           & 0.23                        & 0.31                        & 0.31                                & 0.37                        & 0.37                               & 0.32                               & 0.28                                \\
SynApSe                    & 0.29                           & 0.31                      & 0.32                            & 0.25                           & 0.25                        & 0.28                        & 0.28                                & 0.31                        & 0.38                               & 0.32                               & 0.22                                \\
arioriAveri              & 0.28                           & 0.29                      & 0.32                            & 0.26                           & 0.21                        & 0.27                        & 0.27                                & 0.32                        & 0.34                               & 0.29                               & 0.25                                \\
Data-Wizards                               & 0.26                           & 0.26                      & 0.28                            & 0.23                           & 0.17                        & 0.27                        & 0.25                                & 0.27                        & 0.36                               & 0.29                               & 0.19                                \\
MA-Bros-H & 0.23                           & 0.25                      & 0.28                            & 0.22                           & 0.16                        & 0.23                        & 0.22                                & 0.27                        & 0.3                                & 0.26                               & 0.19                                \\
mitchelldehaven                            & 0.22                           & 0.23                      & 0.24                            & 0.18                           & 0.19                        & 0.22                        & 0.2                                 & 0.22                        & 0.28                               & 0.23                               & 0.19                                \\
SK\_DU                                     & 0.25                           & 0.26                      & 0.27                            & 0.22                           & 0.17                        & 0.25                        & 0.24                                & 0.27                        & 0.34                               & 0.28                               & 0.18                                \\
UPS                                        & 0.26                           & 0.29                      & 0.31                            & 0.25                           & 0.23                        & 0.27                        & 0.28                                & 0.31                        & 0.29                               & 0.27                               & 0.25                                \\
FZI-WIM                              & 0.2                            & 0.22                      & 0.24                            & 0.18                           & 0.12                        & 0.18                        & 0.19                                & 0.21                        & 0.27                               & 0.22                               & 0.15                                \\
KnowComp                                   & 0.2                            & 0.22                      & 0.23                            & 0.18                           & 0.05                        & 0.18                        & 0.19                                & 0.22                        & 0.29                               & 0.23                               & 0.14                                \\
IKR3-UNIMIB                   & 0.23                           & 0.24                      & 0.26                            & 0.19                           & 0.13                        & 0.23                        & 0.21                                & 0.25                        & 0.31                               & 0.25                               & 0.16                                \\
ngetach                                    & 0.21                           & 0.22                      & 0.23                            & 0.18                           & 0.15                        & 0.19                        & 0.2                                 & 0.23                        & 0.24                               & 0.23                               & 0.18                                \\
VGyasi                                     & 0.21                           & 0.22                      & 0.24                            & 0.2                            & 0.11                        & 0.22                        & 0.2                                 & 0.24                        & 0.27                               & 0.24                               & 0.17                                \\
\textit{Baseline}                                   & \textit{0.19}                           & \textit{0.2}                       & \textit{0.23}                            & \textit{0.17}                           & \textit{0.14}                        & \textit{0.19}                        & \textit{0.19}                                & \textit{0.21}                        & \textit{0.24}                               & \textit{0.21}                               & \textit{0.14}                                \\
Factors                                    & 0.19                           & 0.19                      & 0.21                            & 0.16                           & 0.21                        & 0.18                        & 0.16                                & 0.17                        & 0.24                               & 0.2                                & 0.15                                \\
InfinityScalers!                           & 0.11                           & 0.11                      & 0.1                             & 0.08                           & 0.07                        & 0.11                        & 0.1                                 & 0.09                        & 0.22                               & 0.12                               & 0.06                                \\
AYM                                        & 0.13                           & 0.13                      & 0.13                            & 0.1                            & 0.05                        & 0.12                        & 0.11                                & 0.13                        & 0.26                               & 0.14                               & 0.06                                \\ \midrule
Average                                    & 0.25                           & 0.26                      & 0.28                            & 0.22                           & 0.18                        & 0.24                        & 0.24                                & 0.26                        & 0.31                               & 0.26                               & 0.2                \\ \bottomrule                
\end{tabular}}
\caption{Retrieval results in terms of Q+A Hungarian METEOR, broken down according to 1) the document type of the gold evidence, and 2) the number of gold evidence QA pairs for the claim. The overall best performance on retrieval was achieved by Dunamu-ML.}
\label{table:retrieval_breakdown}
\end{table*}

\paragraph{Question Generation \& Retrieval}

Most systems employed an LLM-based question generation strategy. That is, they generated questions or queries, and then retrieved evidence based on those questions. Generating questions, rather than simply searching for the claim, was noted by many top-scoring systems to be essential for good retrieval performance. This supports our hypothesis from \citet{schlichtkrull2023averitec} that question generation (or query expansion~\citep{mao-etal-2021-generation}) is a key avenue for further gains in retrieval.

Question generation was typically implemented using large-scale LLMs, such as GPT-4o or Llama-3.1-70b. Some systems based on smaller model -- HUMANE with Llama-3-8b, UHH with GPT-4o-mini, Data-Wizards with Phi-3-medium, and Papelo with T5 (for the first question only) -- also achieved a high question-only score. This suggests that smaller models can be competitive on search query generation.

Several teams -- Papelo, SynApSe, and IKR3 -- mentioned that they saw benefits from modeling the retrieval task as multi-hop retrieval. That is, instead of retrieving all documents at once, their systems used multiple rounds of retrieval with each round conditional on previous rounds. The benefits of this strategy were also documented in previous FEVER shared tasks, e.g., \citet{malon-2021-team}. Team Papelo further expanded on this strategy, showing that the use of different components at different retrieval steps -- T5 for the first question and GPT-4o for subsequent questions -- yielded higher performance than using a single-question generation model.

As can be seen in Table~\ref{table:detailed_breakdown}, high-performing systems tended to generate and submit a high number of questions. This may be a consequence of our evaluation setup -- there is no brevity penalty (other than documents past the 10th being ignored), so submitting more evidence documents means a higher chance of recalling the gold evidence. 
Several teams also noted that even duplicates of the same question could slightly increase their score. 

We tested this, and observed baseline performance increase by 2 points QA score and 0.5 points \textsc{AVeriTeC} score when including two additional duplicates of each question. There are two reasons this might happen. First, some generated QA pairs may be the best match for multiple gold QA pairs (i.e.\ because they are very long, or because other QA pairs are irrelevant to the claim). Duplicating QA pairs means the generated pair can be matched to multiple gold pairs when computing the Hungarian algorithm, marginally increasing overall performance. Second, Hungarian METEOR is computed by averaging over gold question-answer pairs. If there are more gold pairs than generated pairs, some gold pairs will be \textit{unmatched}. These will receive a score of 0, as the ``matched'' evidence is the empty string, dragging down the average. Effectively, systems are heavily penalised for generating too \textit{few} questions, and may benefit slightly from generating too \textit{many}.

For evidence retrieval, vector-based dense retrieval systems~\citep{karpukhin-etal-2020-dense} were common, along with BM25~\citep{robertson2009bm25}. Several teams -- HUMANE, UHH, SK\_DU -- proposed hybrid systems where coarse retrieval with BM25 was followed by reranking with a vector-based approach. For vector-based retrievers, the \textit{gte}~\citep{li2023gte, zhang2024mgte} family of models were popular, and noted by participants to perform well on the task; this includes Stella\footnote{\url{https://huggingface.co/dunzhang/stella\_en\_400M\_v5}}, an MRL~\citep{kusupati2022matryoshka} approach based on \textit{gte}. Several teams noted that their \textit{gte}- or Stella-based retrievers were chosen as they, at the time of the competition, were top performers on the MTEB~\citep{muennighoff-etal-2023-mteb} leaderboard.

The overall best performing retrieval system was Dunamu-ML, closely followed by HUMANE. In Table~\ref{table:retrieval_breakdown}, we break down performance on retrieval according to which document type the \textit{gold} evidence originated from. We see that Dunamu-ML do have top performance on PDFs and videos (for which they added a custom scraper), but tie respectively with HUMANE and TUDA\_MAI on these categories. On the other hand, Dunamu-ML perform better than other systems on tabular and image evidence, while HUMANE is the top performer on Metadata, Audio, and ``Other'' evidence (used by participants mostly for social media posts, as well to link to external web tools, such as a calculator in support of numerical reasoning).

In Table~\ref{table:retrieval_breakdown}, we also break down retrieval performance by the number of gold evidence question-answer pairs per claim. HUMANE performs the best on claims with only one gold document, narrowly followed by Dunamu-ML. As the number of claims increases, Dunamu-ML takes the lead. With an average of 2.74 questions per claim in the test set, this may explain why Dunamu-ML achieved the overall highest retrieval performance. 

\begin{table*}[t]
    \centering
    \begin{tabular}{l|rrrrr|rrrr|r} \toprule
Team name                     & QV & N & E/P & C & PS & S & R & NEE & CE/C & Avg. \# Docs \\ \midrule
TUDA\_MAI      & \textbf{0.64}               & \textbf{0.58}            & \textbf{0.64}                 & \textbf{0.64}         & \textbf{0.58}               & 0.64      & \textbf{0.73}    & 0.12                & 0.19                               & 9.3            \\
HUMANE                & 0.59               & 0.57            & 0.58                 & 0.55         & 0.46               & \textbf{0.76}      & 0.62    & 0.01                & 0.12                               & 10.0           \\
CTU AIC                       & 0.57               & 0.49            & 0.51                 & 0.52         & 0.38               & 0.58      & 0.58    & 0.1                 & 0.01                               & 9.89           \\
Dunamu-ml                     & 0.44               & 0.49            & 0.5                  & 0.55         & 0.4                & 0.69      & 0.5     & \textbf{0.31}                & 0.12                               & 12.41          \\
Papelo                        & 0.51               & 0.38            & 0.5                  & 0.51         & 0.45               & 0.45      & 0.59    & 0.0                 & 0.0                                & 9.95           \\
UHH                           & 0.46               & 0.43            & 0.46                 & 0.48         & 0.39               & 0.47      & 0.54    & 0.0                 & 0.0                                & 10.0           \\
SynApSe                       & 0.45               & 0.39            & 0.43                 & 0.43         & 0.36               & 0.42      & 0.5     & 0.02                & \textbf{0.21}                               & 4.26           \\
arioriAveri & 0.44               & 0.37            & 0.39                 & 0.4          & 0.29               & 0.45      & 0.44    & 0.09                & 0.06                               & 8.98           \\
Data-Wizards                  & 0.37               & 0.3             & 0.34                 & 0.32         & 0.29               & 0.44      & 0.36    & 0.05                & 0.04                               & 3.0            \\
MA-Bros-H                     & 0.29               & 0.3             & 0.26                 & 0.25         & 0.19               & 0.4       & 0.27    & 0.08                & 0.0                                & 3.74           \\
mitchelldehaven               & 0.24               & 0.26            & 0.25                 & 0.25         & 0.16               & 0.4       & 0.25    & 0.0                 & 0.0                                & 5.0            \\
SK\_DU                        & 0.27               & 0.3             & 0.21                 & 0.15         & 0.14               & 0.36      & 0.22    & 0.01                & 0.11                               & 3.0            \\
UPS                           & 0.29               & 0.18            & 0.22                 & 0.2          & 0.21               & 0.17      & 0.24    & 0.08                & 0.14                               & 10.0           \\
FZI-WIM                       & 0.21               & 0.25            & 0.18                 & 0.16         & 0.21               & 0.31      & 0.18    & 0.12                & 0.02                               & 2.52           \\
KnowComp                      & 0.16               & 0.19            & 0.19                 & 0.15         & 0.13               & 0.27      & 0.19    & 0.0                 & 0.01                               & 2.55           \\
IKR3-UNIMIB            & 0.21               & 0.22            & 0.17                 & 0.17         & 0.15               & 0.28      & 0.19    & 0.01                & 0.05                               & 3.0            \\
ngetach                       & 0.16               & 0.13            & 0.14                 & 0.17         & 0.09               & 0.0       & 0.22    & 0.0                 & 0.0                                & 4.25           \\
VGyasi                        & 0.16               & 0.11            & 0.13                 & 0.11         & 0.10                & 0.1       & 0.12    & 0.22                & 0.03                               & 3.46           \\
\textit{Baseline}                      & \textit{0.14}               & \textit{0.16}            & \textit{0.11}                 & \textit{0.10}          & \textit{0.06}               & \textit{0.17}      & \textit{0.12}    & \textit{0.0}                 & \textit{0.04}                               & \textit{3.0 }           \\
InfinityScalers!              & 0.04               & 0.10             & 0.09                 & 0.08         & 0.08               & 0.24      & 0.04    & 0.04                & 0.10                                & 3.52           \\
AYM                           & 0.07               & 0.06            & 0.06                 & 0.03         & 0.10                & 0.11      & 0.05    & 0.0                 & 0.0                                & 1.0            \\
Factors                       & 0.04               & 0.05            & 0.05                 & 0.05         & 0.04               & 0.13      & 0.03    & 0.04                & 0.01                               & 1.0            \\ \midrule
Average                       & 0.31               & 0.29            & 0.29                 & 0.29         & 0.24               & 0.36      & 0.32    & 0.06                & 0.06                               & 5.63          \\ \bottomrule
\end{tabular}
    \caption{We compute separate results based on claim type (QV = Quote Verification, N = Numerical, E/P = Event/Property, C = Causal, PS = Position Statement). We also compute results separated by gold verdict (S = Supported, R = Refuted, NEE = Not Enough Evidence, CE/C = Conflicting Evidence / Cherrypicking). Finally, we report the average number of evidence documents submitted per claim. We note that if a team submitted more than 10 documents for a claim, only the first 10 were used to compute retrieval scores for evaluation.}
    \label{table:detailed_breakdown}
\end{table*}

\paragraph{Veracity Prediction}

Veracity prediction was also dominated by LLM-based approaches, including GPT-4o, Llama 3.1, and Mixtral. 
Teams HUMANE and SynApSe note that some fine-tuning was necessary for good performance on veracity prediction. Various teams saw improvements both from full fine-tuning of all the weights, and from fine-tuning with LORA~\citep{hu2022lora}. Interestingly, one team -- Papelo -- chose to prevent their veracity prediction system from predicting Not Enough Evidence and Conflicting Evidence, arguing that their prompting-based model too frequently chose these rarer labels. This may explain why  calibration was especially helpful for this task.

We note that top-scoring systems tended to use very large models for veracity prediction, such as GPT-4o, Llama-3.1-70b, or Mixtral-8x7b. The superior reasoning capabilities of these cutting-edge models appear especially critical to this stage of the pipeline, unlike for question generation.  

\paragraph{Types \& Verdicts}

In Table~\ref{table:detailed_breakdown}, we provide a detailed breakdown of the results based on claim type (quote verification, numerical claims, event/property claims, causal claims, position statements) and verdict (supported, refuted, conflicting evidence/cherrypicking, not enough evidence). For each category, we report \textsc{AVeriTeC} scores on the corresponding subset of the test set.

Systems performed slightly better on quote verification, slightly worse on position statements, and approximately equally well on other claims. This is interesting, as quote verification and position statements are relatively similar tasks. In the former, systems must verify if a person has uttered a quote verbatim; in the latter, systems must verify if a person or organisation holds a specific position (e.g., supporting a policy), but not necessarily verbatim. Verifying position statements often required abductive reasoning, which LLMs are known to struggle with~\citep{dougrezlewis2024assessingreasoning}.

Among the top performing systems, performance is frequently lower on numerical statements (along with position statements) compared to other claims. This suggests that the gap is smaller for numerical reasoning than other forms of reasoning. As top performers often use very large LLMs, that 
is suggestive of the type of reasoning gains accomplished by scaling up these models.

In terms of performance across the different labels, there is significant variation. First, systems often have different calibration to predict supported versus refuted claims. As refuted claims dominate (making up approximately two-thirds of the dataset), this yields a significant advantage for some participants.
We note that a common strategy among participants was to ignore the rarer veracity labels -- not enough evidence, and conflicting evidence. As mentioned e.g.\ by team Papelo in their system description paper, large language models tend to overpredict these rarer classes. Nevertheless, many top performers, including the winning system, made significant gains on these classes. 

\paragraph{Quality Controls on Test Submissions}
To ensure the reliability of submitted systems, we conducted quality control on our submissions.
Here, \emph{reliability} refers to the evidence (QA pairs) being grounded and supported by their retrieved documents. Typically, participants returned answers generated based on retrieved documents; although some systems generated answers e.g. with an LLM, and subsequently matched the answer to a ``backing document''.

We first used an automatic method to evaluate the entailment between the answers and the retrieved documents. 
Specifically, we applied a DeBERTa-large-based NLI model~\citep{he2020deberta}\footnote{\url{https://huggingface.co/MoritzLaurer/DeBERTa-v3-large-mnli-fever-anli-ling-wanli}, which demonstrates the best performance on NLI tasks amongst Hugging Face models.} on all submissions, taking each answer as hypothesis and its corresponding document as premise.
Generally, we find that most teams see a small proportion of entailment labels and a large proportion of neutral labels (~80\%).
This can be because the NLI model cannot perform well on out-of-distribution data in a zero-shot setting, in particular when the retrieved document is much longer than the standard NLI premise (e.g., the average document length in words in TUDA\_MAI's submission is over 4,000, while it is around 50 in ANLI~\citep{mishra-etal-2021-looking}).


Therefore, we further investigated submissions via manual evaluation. In particular, we focused on instances which the NLI model identified as either \emph{neutral} or \emph{contradiction}, and on the top-4 performing systems (i.e.: TUDA\_MAI, HUMANE, CTU AIC and Dunamu-ml).
We randomly selected 20 neutral or contradicting instances from each submission, and then performed human evaluation.
Given an instance with its corresponding QA pairs and retrieved documents, we identified whether the answers were entailed by the retrieved documents.

Generally, we found that all systems were mostly reliable, with the evidence they generate being supported by the retrieved documents. All answers from TUDA\_MAI were extractive from source documents and thus entailed. 
The answers from the other three systems were more abstractive. 
Although the answers can contain some hallucination (e.g., generating answers that contradict the retrieved documents by mistake), our manual evaluation found the answers were mostly (HUMANE: 19/20; CTU AIC: 17/20; Dunamu-ml: 12/20) entailed by their associated documents. Errors were typically due to mistakes by the question-answering components, such as taking a snippet from the associated document out of context. 
Thus, we conclude that the systems evaluated were reliable and find relevant documents that provide useful information for predicting veracity.

\section{Human Evaluation of Evidence}

Following the approach taken in the first FEVER shared task \citep{thorne-etal-2018-fact}, we conducted human evaluation of the evidence retrieved by the systems participating in the shared task, motivated by two concerns. First, the incompleteness of the gold evidence annotation, since it is often the case that adequate evidence to determine the verdict for a claim can be found in multiple webpages, as shown in the inter-annotation agreement study of \citet{schlichtkrull2023averitec}. Second, the inaccuracies of automatic evaluation metrics of textual evaluation, especially in the case of token-matching metrics such as METEOR \citep{banerjee-lavie-2005-meteor} used here, but also of more recent neural ones such as FactScore \cite{min-etal-2023-factscore}.
Thus we can gain a deeper understanding of the quality of the retrieved evidence, and assess how well the \textsc{AVeriTeC} scores assigned to the retrieved evidence aligns with human judgements.

\paragraph{Evaluation Process}

We conducted human evaluation in collaboration with the participating teams. Sixteen top-performing teams were invited to participate in the evaluation. However, teams Dunamu-ml, mitchelldehaven, and KnowComp did not take part. Each of the remaining thirteen participating teams manually evaluated thirty evidence samples from other participants. Out of these, five were gold-labeled, which were included to assist in the post-processing of the collected annotations and to assess their quality. The evidence samples were randomly selected and evenly distributed across all submitted systems, representing both high- and low-scoring systems, as shown in Table~\ref{table:detailed_breakdown}.

Figures in Appendix~\ref{appendix:human_eval} depict the evaluation form and the instructions provided to human annotators during evaluation. As a first step, we asked annotators to assess whether ``at least some part of the evidence'' was ``non-empty, understandable, and related to the claim.'' If so, it was considered eligible for further rating. In addition to assigning a verdict label, we asked annotators to rate retrieved evidence in comparison to provided reference evidence\footnote{We provide the exact instruction for rating each criteria in the appendix.}. Annotators rated the evidence on a scale from 1 to 5 across five dimensions:


\noindent \textbf{$(1)$ Coverage}: Measures how much of the reference evidence is covered by the predicted evidence, ensuring that the content, meaning, entities, and other key elements of the reference are fully represented in the retrieved evidence.

\noindent \textbf{$(2)$ Coherence}: Captures whether the retrieved evidence is coherent, i.e., if all sentences are connected sensibly and the evidence makes sense as a whole.

\noindent \textbf{$(3)$ Repetition}: Evaluates whether the retrieved evidence exhibits repetition of its content. 

\noindent \textbf{$(4)$ Consistency}: Assesses whether the retrieved evidence is semantically consistent and does not contain conflicting information. Unlike coherence, which focuses on how well the information is structured, consistency evaluates whether the arguments presented in the evidence for or against a claim are sound and aligned.

\noindent \textbf{$(5)$ Relevance}: Measures how relevant the retrieved evidence is to the content of the claim.

\paragraph{Insights Gained}

The annotation process resulted in a total of $389$ annotations. After filtering out evidence samples that were labeled by evaluators as entirely empty ($1\%$), not understandable ($1.8\%$), or completely irrelevant to the given claim ($9.4\%$), we were left with $344$ valid annotations. Among these, $66$ annotations corresponded to gold-labeled samples. Excluding the gold-labeled samples, resulted in a final set of $278$ evidence annotations.

Before labeling the system-retrieved evidence, participants were first asked to label the verdict of the retrieved evidence. Table~\ref{tab:evidence_summary} provides an overview of the matching between system-predicted labels (columns) and human-labeled verdicts (rows). While human annotators generally agreed with evidence labeled as refuted or supported, there was less overlap for evidence labeled as NEE and CE/C by the submitted systems.

\begin{table}[t]
    \centering
    \small
    \begin{tabular}{c|c|c|c|c}
    \toprule
    \textbf{Label/Pred} & \textbf{CE/C} & \textbf{NEE} & \textbf{Refuted} & \textbf{Supported} \\
    \midrule
     \textbf{CE/C} & 35.7 & 3.6 & 53.6 & 7.1 \\
     \textbf{NEE} & 5.9 & 22.1 & 60.3 & 11.8 \\
     \textbf{Refuted} & 3.9 & 4.9 & 85.4 & 5.8 \\
     \textbf{Supported} & 7.6 & 0 & 16.5 & 76.0 \\
    \bottomrule
    \end{tabular}
    \caption{Overview of verdict \textbf{label}led by human evaluators (rows) versus system \textbf{pred}ictions (columns).}
    \label{tab:evidence_summary}
\end{table}

Analyzing human judgments across the five evaluated dimensions (see Table~\ref{tab:human_eval_scores}), we find that the majority of predicted evidence was labeled as very coherent, consistent, relevant, and containing limited repetition. However, in the dimension of semantic coverage, approximately $15\%$ of the evidence received a rating of $0$, indicating that ``the predicted evidence covers none of the reference evidence.'' Additionally, around $20\%$ received a rating of $1$, meaning that ``very little of the reference evidence is covered.'' 
This does not necessarily mean that the evidence is false -- low coverage can also occur if the retrieved evidence uses different information, arguments, or sources than the reference evidence. Ideally, we aim for an 
evidence evaluation 
that can fairly assess evidence even when it differs from the reference and has low coverage.

\begin{table*}[t]
    \centering
    \small
    \begin{tabular}{c|cc|cc|cc|cc|cc}
    \toprule
     \textbf{Rating} & \textbf{COV} & \textbf{COV \%} & \textbf{COH} & \textbf{COH \%} & \textbf{REP} & \textbf{REP \%} & \textbf{CON} & \textbf{CON \%} & \textbf{REL} & \textbf{REL \%} \\
    \midrule
     1& 42 & 15.16 & 4 & 1.44 & 23 & 8.27 & 6 & 2.17 & 4 & 1.44 \\
     2& 59 & 21.30 & 42 & 15.11 & 51 & 18.35 & 35 & 12.64 & 26 & 9.35 \\
     3& 59 & 21.30 & 64 & 23.02 & 61 & 21.94 & 57 & 20.58 & 51 & 18.35 \\
     4& 71 & 25.63 & 81 & 29.14 & 71 & 25.54 & 82 & 29.60 & 83 & 29.86 \\
     5& 46 & 16.61 & 87 & 31.29 & 72 & 25.90 & 97 & 35.02 & 114 & 41.01 \\
    \bottomrule
    \end{tabular}
    \caption{Overview of ratings for Semantic \textbf{Cov}erage, \textbf{Coh}erence, \textbf{Rep}etition, \textbf{Con}sistency, and \textbf{Rel}evance. For each evaluation dimension, the first column depicts the absolute number of annotations for a specific score (from 1 to 5) and the second column the percentages.}
    \label{tab:human_eval_scores}
\end{table*}

To assess the relationship between human scoring and the Hungarian METEOR (see Sec~\ref{ssec:evaluation}), we computed both the Spearman correlation coefficient ($\rho$~\citep{Spearman1987}) and the Pearson correlation coefficient ($r$~\citep{Pearson1896}) as shown in Table~\ref{tab:correlation_metrics}. Correlations were calculated using both the entire evidence text and the question text only. In both cases, we observed a low correlation between the Hungarian Meteor and the assessed dimensions, with the highest correlation seen in the category of ``repetition'' (see Table~\ref{tab:correlation_metrics}).
While the results show a similar ranking of participating systems compared to human evaluations on the subset, further work is needed to develop scoring methods that align more closely with human assessments of evidence. With that said, overall, the top-ranked teams (based on \textsc{AVeriTeC} score) also perform well on human evaluation, while the lower-ranked teams remain similarly positioned, with only minor shifts in their order.\footnote{See Table~\ref{tab:human_eval_scores} in the appendix.} It is important to note that this evaluation was solely based on a small sample of system predictions, and that the results should therefore be taken with a grain of salt.

\begin{table}[t]
    \centering
    \begin{tabular}{lcc}
    \toprule
     \textbf{Dimension} & \textbf{$\rho$} & \textbf{$r$} \\
    \midrule
     Coverage & .005 & -.024 \\
     Coherence & .076 & .057 \\
     Repetition & .117 & .025 \\
     Consistency & .039 & .024 \\
     Relevance & .008 & .003 \\
    \bottomrule
    \end{tabular}
    \caption{Correlation between Q + A scores (Hungarian METEOR) and human-rated subset of evidence. We calculate correlation using the Spearman ($\rho$) and Pearson ($r$) correlation coefficients.}
    \label{tab:correlation_metrics}
\end{table}

Human evaluation of evidence predictions offers valuable insights into the limitations of the \textsc{AVeriTeC} score, and suggests directions for future research. A notable observation is the discrepancy between human evaluation and the \textsc{AVeriTeC} score for some of the highest-ranked samples, such as the examples provided in Table~\ref{tab:evidence_evaluation_high} in the appendix. For instance, in row three, the predicted evidence directly contradicts the reference evidence by providing different numbers, yet it receives a high \textsc{AVeriTeC} score due to similar wording. Similarly, for the first two rows in Table~\ref{tab:evidence_evaluation_high}, the semantic coverage score is rated with the second lowest score 1, whereas the average score across all examples is 3, indicating misalignment between the predicted and reference evidence.

Certain low-ranked examples highlight different challenges (see Table~\ref{tab:evidence_evaluation_low}). For example, the predicted evidence in the first row received a low \textsc{AVeriTeC} score despite receiving the highest score of 5 across all categories in human evaluation. 
Despite both sets of evidence reaching the same conclusion, the large disparity in answer length and wording leads to a much lower \textsc{AVeriTeC} score. The example in the second row, also ranks low according to \textsc{AVeriTeC} score, even though it scores high in all categories except for coverage, where it scores 3. Here, both the reference and predicted evidence reach the same verdict, but the predicted evidence supports the claim with different information and wording, resulting in low semantic coverage and a low \textsc{AVeriTeC} score. 

\section{Lessons Learned}

Providing a knowledge store rather than requiring participants to rely on a search engine API made the task more accessible. Given the cost of API access, this allowed substantial analysis and work by participants on retrieval. We note that most submissions -- 13 of 16 system description papers -- used the knowledge store. Nevertheless, because of the size of the knowledge store and the inclusion of distractor documents, the knowledge store did not trivialise the task, and systems relying on search remain competitive and provide unique advantages. Several participants, such as team FZI-WIM, commented on how the two are complementary, and suggested hybrid systems using \textit{both} as a potentially fruitful extension of their systems.

\textsc{AVeriTeC} presupposes a strong focus on evidence retrieval. The overall score, as in FEVER~\citep{thorne-etal-2018-fever}, is determined \textit{both} by retrieval performance \textit{and} by veracity prediction performance. In the \textsc{AVeriTeC} shared task, participant systems innovated across the pipeline, and all of the top-scoring systems suggest improvements to multiple subtasks of fact-checking. 



\begin{table}[t]
    \centering
    \begin{tabular}{lrr}
\toprule
Team name                     & 0-1000 & 1000-2215 \\ \midrule
TUDA\_MAI      & \textbf{0.61}        & \textbf{0.64}                           \\
HUMANE                 & 0.55        & 0.58                           \\
CTU AIC                       & 0.45        & 0.55                           \\
Dunamu-ml                     & 0.5         & 0.5                            \\
Papelo                        & 0.49        & 0.46                           \\
UHH                           & 0.41        & 0.48                           \\
SynApSe                       & 0.41        & 0.43                           \\
arioriAveri & 0.35        & 0.42                           \\
Data-Wizards                  & 0.32        & 0.34                           \\
MA-Bros-H                     & 0.22        & 0.31                           \\
mitchelldehaven               & 0.22        & 0.27                           \\
SK\_DU                        & 0.2         & 0.25                           \\
UPS                           & 0.15        & 0.25                           \\
FZI-WIM                       & 0.19        & 0.2                            \\
KnowComp                      & 0.19        & 0.18                           \\
IKR3-UNIMIB            & 0.16        & 0.2                            \\
ngetach                       & 0.12        & 0.16                           \\
VGyasi                        & 0.12        & 0.12                           \\
\textit{Baseline}                      & \textit{0.11}        & \textit{0.12}                           \\ 
InfinityScalers!              & 0.1         & 0.07                           \\
AYM                           & 0.06        & 0.06                           \\
Factors                       & 0.06        & 0.04                           \\ \midrule
Average                       & 0.27        & 0.3                           \\ \bottomrule
\end{tabular}
    \caption{\textsc{AVeriTeC} scores for different subsections of the dataset. We compute results for the initial test set of 1000 examples collected by~\citet{schlichtkrull2023averitec}, and for the additional 1215 test examples collected for this shared task.}
    \label{table:performance_by_part}
\end{table}

When submitting test set predictions, we required participants to include a field (``\textit{scraped\_text}'') for each piece of evidence in their submission, corresponding to the webpage providing backing for that piece of evidence. This enabled us to carry out manual and automatic quality control evaluation verifying that systems do indeed ground their evidence in external sources (see Section~\ref{sec:results}). This enabled us to detect, for example, if some systems were hallucinating evidence; we did not see any evidence of hallucinated evidence, but we consider guardrails against this crucial. Unfortunately, the inclusion of this field made some submissions substantial in size, as entire webpages were included -- up to 2.3gb for the largest submission. Our submission portal, eval.ai, was not able to handle these large files, blocking the portal for all participants during the last few days of the competition. We extended the deadline to compensate.

The scraper we used for the knowledge store (same as in \citet{schlichtkrull2023averitec}) to retrieve evidence turned out to be a significant weakness. As some participants noticed, many knowledge store documents are empty. The submission with the best retrieval performance, Dunamu-ml, used a custom scraper, and may have derived significant gains from that choice. We suggest that this may be an interesting area for further research.


During the competition, we identified an issue with the knowledge store data for the last 1215 test examples. Due to an error with date formats, for some claims, web pages published after the claim were included in the knowledge store. This included fact-checking articles, as also mentioned by CTU AIC in their system description paper. As the first 1000 examples were not affected, we computed performance on the first 1000 and last 1215 test examples separately -- see Table~\ref{table:performance_by_part}.

As can be seen, the ranking of participants on the two splits is roughly the same -- and, indeed, roughly the same as for the entire test set. The second half \textit{was} easier, and many systems perform slightly better there. Somewhat surprisingly, some systems which relied on Google search -- specifically, SynApSe -- \textit{also} saw a performance gain when measured only on the second split. As such, we do not believe this issue majorly impacted any subset of participants, such as those not relying on the knowledge store. We release an updated knowledge store along with our shared task paper, accessible at \url{https://fever.ai/dataset/averitec.html}. We have re-compiled the knowledge store with the correct date cutoff, and removed any fact-checking articles that snuck through from the evidence base.

\section{Conclusions \& Future Work}

The \textsc{AVeriTeC} shared task attracted submissions from 21 teams, 18 of which outperformed our baseline. The leaderboard was dominated by systems relying on large language models, especially GPT-4o; nevertheless, especially for question generation and retrieval, smaller models -- such as LLama-3-8b -- also achieved top performance. The winner of the shared task was team TUDA\_MAI, which achieved an \textsc{AVeriTeC}-score of 63\%. In this paper we have analysed the shared task, highlighting aspects of the 16 submitted system description papers, as well as key takeaways from the shared task itself.

The strong performance of the participating teams establishes a firm foundation for automating aspects of real-world fact-checking. The results furthermore indicate clear directions for future work. First, most participating systems -- especially for veracity prediction -- relied on very large models, such as GPT-4. Further, many of these are blackbox models. These models may be prohibitively expensive for some real-world use cases, e.g., assisting smaller fact-checking organisations~\citep{schlichtkrull-etal-2023-intended}. Given that, we suggest that getting smaller, more efficient models to reach the performance of their larger counterparts may be a fruitful direction for further research. Similarly, we note that performance for most top-scoring systems was much higher on supported and refuted claims, compared to conflicting evidence and not enough evidence. We suggest that leveling this gap is another clear avenue for future improvements.

\section{Limitations \& Ethics}

The datasets and models described in this paper are not intended for truth-telling, e.g.\ for the design of fully automated content moderation systems. The evidence selection and veracity labels provided in the \textsc{AVeriTeC} dataset relate only to the evidence recovered by annotators, and as such are subject to the biases of annotators and journalists. Participant systems, which sought to maximize performance on \textsc{AVeriTeC}, may replicate those biases. We furthermore note that shared task leaderboards are a limited representation of real-world task needs, not the least because the test set is static. Acting on veracity estimates arrived at through biased means, including automatically produced ranking decisions for evidence retrieval, risks causing epistemic harm~\citep{schlichtkrull-etal-2023-intended}.

\section*{Acknowledgments}
Michael, Yulong, Chenxi, Zhenyun, and Andreas received funding from the European Research Council (ERC) under the European Union’s Horizon 2020 Research and Innovation programme grant AVeriTeC (Grant agreement No. 865958). Rui is funded by a grant from the Alan Turing Institute and DSO National Laboratories (Singapore). Rami Aly was supported by the Engineering and Physical Sciences Research Council Doctoral Training Partnership (EPSRC). 
The annotation of the new test set was conducted by a donation from Google.

\bibliography{anthology,custom}

\appendix

\section{Search Queries for Knowledge Store Generation}
\label{appendix:search_query_prompts}

When creating the knowledge stores for the train, development, and test set, we used a series of search query generation strategies. An overview can be seen in Table~\ref{table:knowledge_store_prompts}. We note that some of these rely on information not available normally to participants, such as the gold question-answer pairs. We note that, despite this, systems not relying on the knowledge store, such as Papelo, were competitive.

\section{Human Evaluation}
\label{appendix:human_eval}

We carried out human evaluation of the submitted test set predictions. Below in Figures~\ref{fig:test1}-\ref{fig:test8}, we include screenshots of the interface used by annotators. We also include, in Tables~\ref{tab:evidence_evaluation_high} and ~\ref{tab:evidence_evaluation_low}, instructive examples from the human evaluation.

\begin{table}[t]
    \centering
    \begin{tabular}{lc}
    \toprule
     \textbf{Source} & \textbf{Score Coverage} \\
    \midrule
     CTU AIC & 4.1 \\
     TUDA\_MAI & 4.1 \\
     SynApSe & 3.8 \\
     Dunamu-ML & 3.5 \\
     MA-Bros-H & 3.4 \\
     Factors & 3.3 \\
     Data-Wizards & 3.2 \\
     UHH & 3.2 \\
     mitchelldehaven & 3.1 \\
     SK\_ DU & 3.1 \\
     IKR3-UNIMIB & 3.1 \\
     FZI-WIM & 2.9 \\
     InfinityScalers! & 2.9 \\
     arioriAveri & 2.9 \\
     HUMANE & 2.8 \\
     Papelo & 2.8 \\
     KnowComp & 2.8 \\
     UPS & 2.4 \\
     VGyasi & 2.3 \\
     AYM & 2.3 \\
     ngetach & 2.0 \\
    \bottomrule
    \end{tabular}
    \caption{Average scores assigned to evidence samples from different participating teams for the semantic coverage category, based on human evaluation.}
    \label{tab:human_eval_scores}
\end{table}

\begin{table*}[ht]
    \centering
    \begin{tabular}{p{0.3\linewidth} p{0.63\linewidth}} \toprule
Query type & Description \\\midrule
Generated questions & \textit{Questions are generated with gpt-3.5-turbo based on the claim. Three claim-question pairs from the training set are used as in-context examples}. \\
Generated background queries & \textit{Queries are generated with gpt-3.5-turbo based on the claim. The prompt focuses on background information, such as details about entities in the claim. Three manually constructed claim-query pairs are used as in-context examples}.\\
Generated provenance queries & \textit{Queries are generated with gpt-3.5-turbo based on the claim. The prompt focuses on information necessary to establish provenance, such as whether the claim source is a satire site. Three manually constructed claim-query pairs are used as in-context examples}. \\
Claim named entities & \textit{Named entities from the claim are extracted and used as search queries. One query for each entity is constructed, along with one query containing all entities.} \\
Most similar gold evidence & \textit{The most similar paragraph in the gold evidence document is selected using BM25, and used as a search query.} \\
Gold URL generated questions &  \textit{Queries are generated with gpt-3.5-turbo based on the URL of the gold evidence. The prompt tried to generate questions that would retrieve the URL in question. Three manually constructed URL-query pairs are used as in-context examples}.\\
Different event same entity &  \textit{Queries are generated with gpt-3.5-turbo based on the named entities in the claim. The prompt focuses on different events involving some of the same entities. Results are used as distractors to make the retrieval task harder}. \\
Similar entities & \textit{Queries are generated with gpt-3.5-turbo based on the claim. The prompt replaces entities in the claim with other similar entities, such as changing one city to another. Results are used as distractors to make the retrieval task harder}. \\
Gold questions & \textit{Gold questions used verbatim as search queries.} \\
Claim + gold question & \textit{Gold questions used verbatim as search queries. The claim is prepended, processed as in \citet{schlichtkrull2023averitec}.} \\
Rephrased gold questions & \textit{Gold questions are rephrased using gpt-3.5-turbo, and then input as search queries.} \\ 
Gold answers & \textit{Gold questions used verbatim as search queries.} \\
Rephrased gold answers & \textit{Gold answers are rephrased using gpt-3.5-turbo, and then input as search queries.} \\\bottomrule
    \end{tabular}
    \caption{Queries input to the Google Search API for each claim in order to build the knowledge store. Following \citet{schlichtkrull2023averitec}, we restrict search results to documents published before the claim. For each claim, we also extend the knowledge store with the corresponding gold evidence documents.}
    \label{table:knowledge_store_prompts}
\end{table*}

\begin{figure*}[ht]
    \centering
    \includegraphics[scale=0.6]{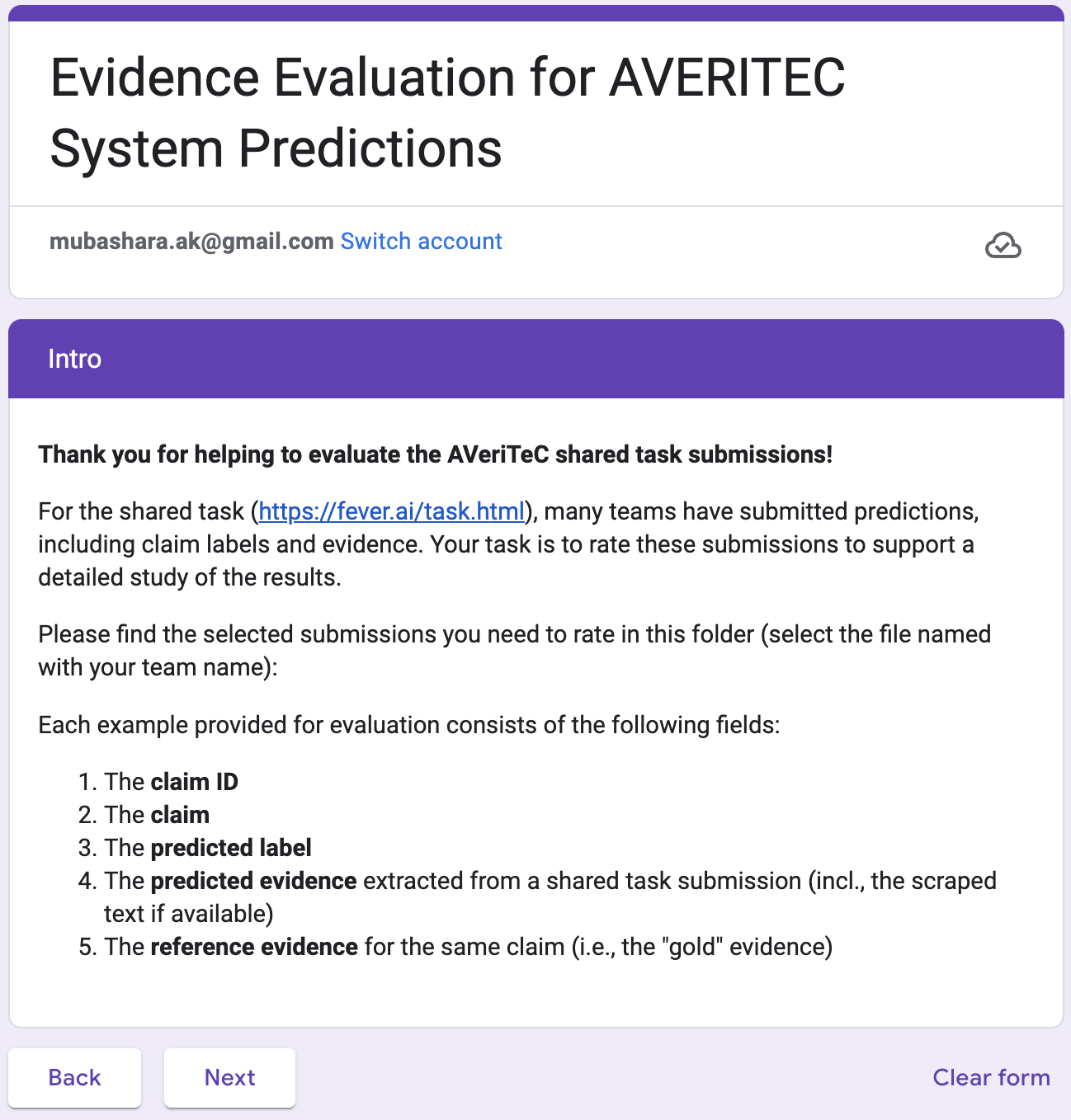}
    \caption{Platform for human evaluation of retrieved evidence from participating systems.}
    \label{fig:test1}
\end{figure*}

\begin{figure*}[ht]
    \centering
    \includegraphics[scale=0.75]{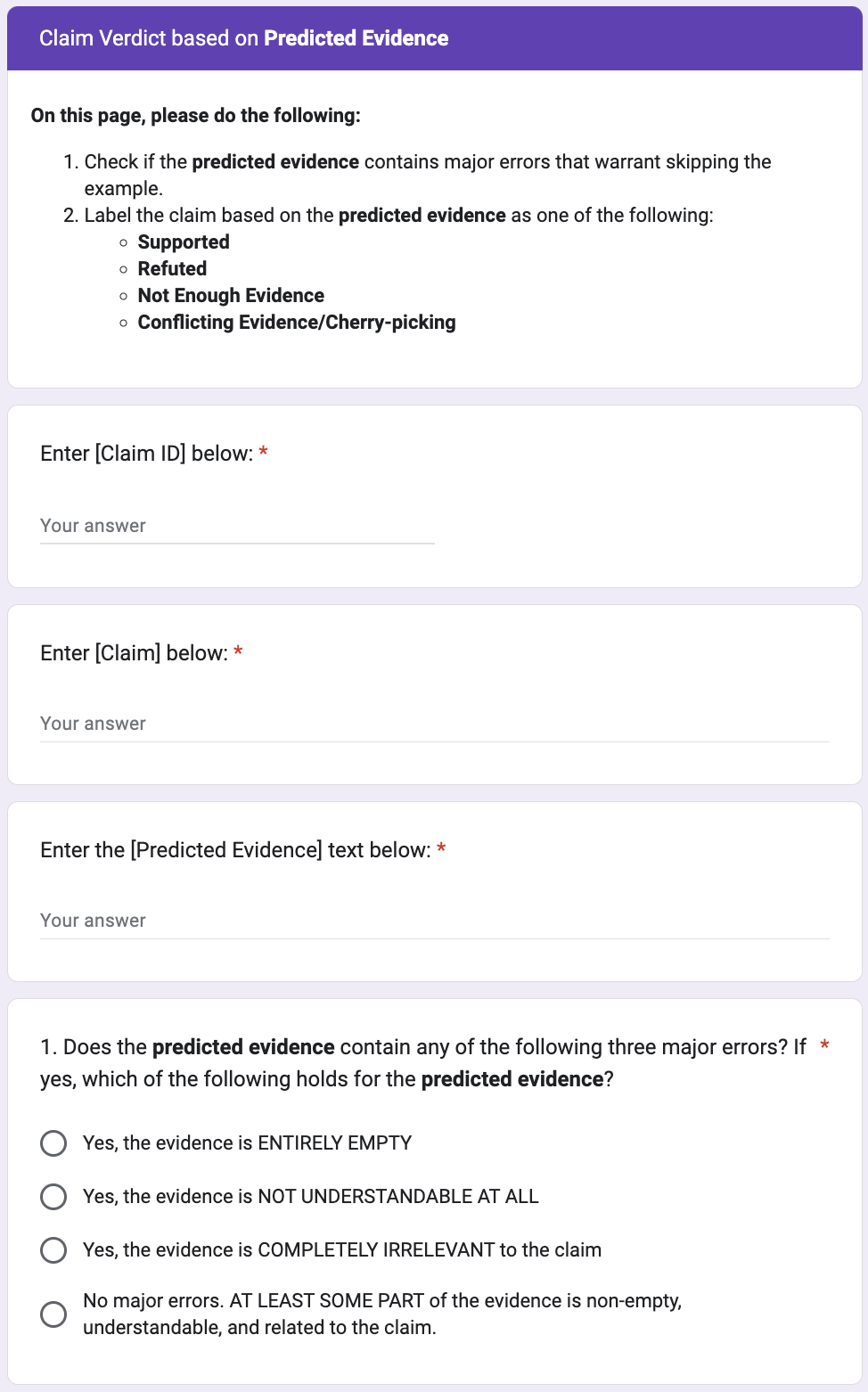}
    \caption{Platform for human evaluation of retrieved evidence from participating systems.}
    \label{fig:test2}
\end{figure*}

\begin{figure*}[ht]
    \centering
    \includegraphics[scale=0.75]{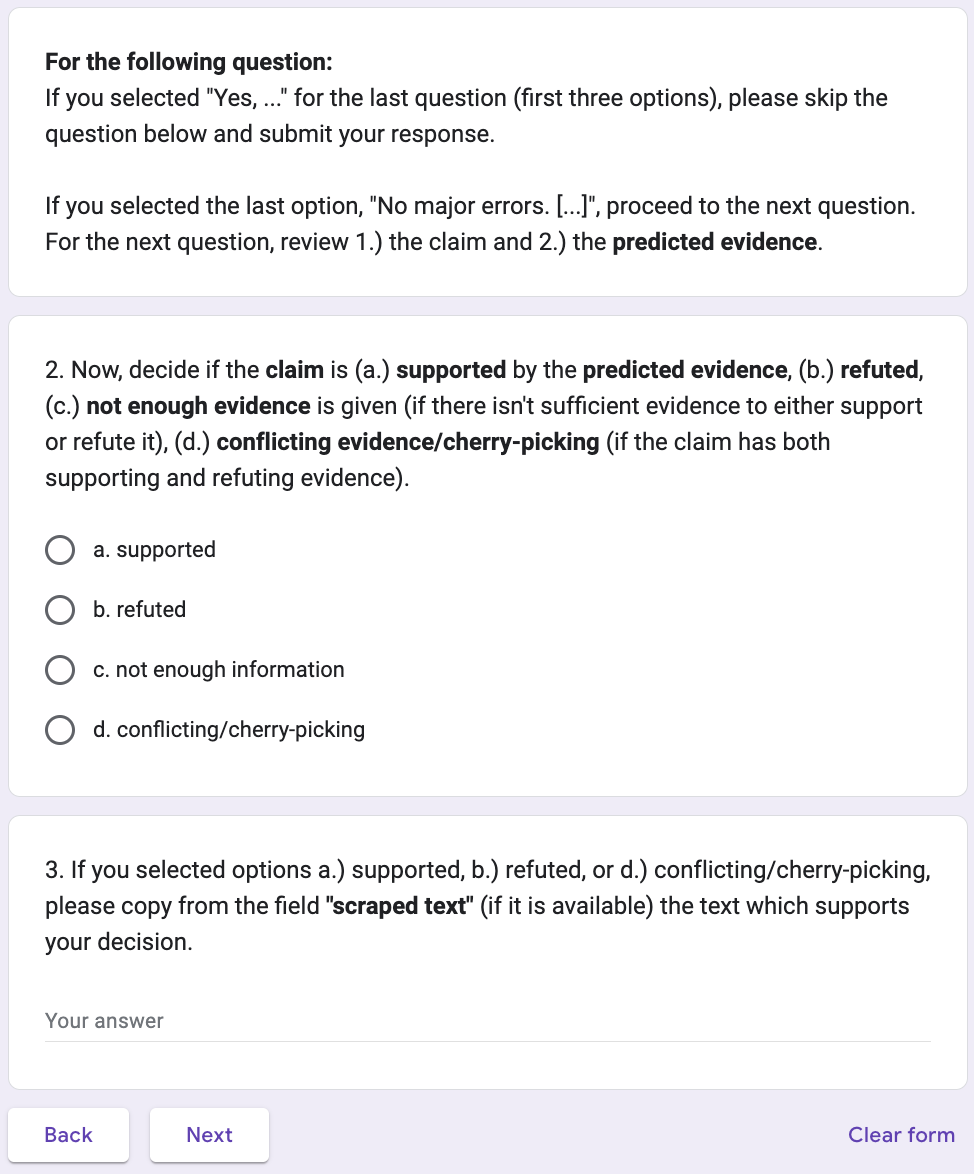}
    \caption{Platform for human evaluation of retrieved evidence from participating systems.}
    \label{fig:test3}
\end{figure*}

\begin{figure*}[ht]
    \centering
    \includegraphics[scale=0.6]{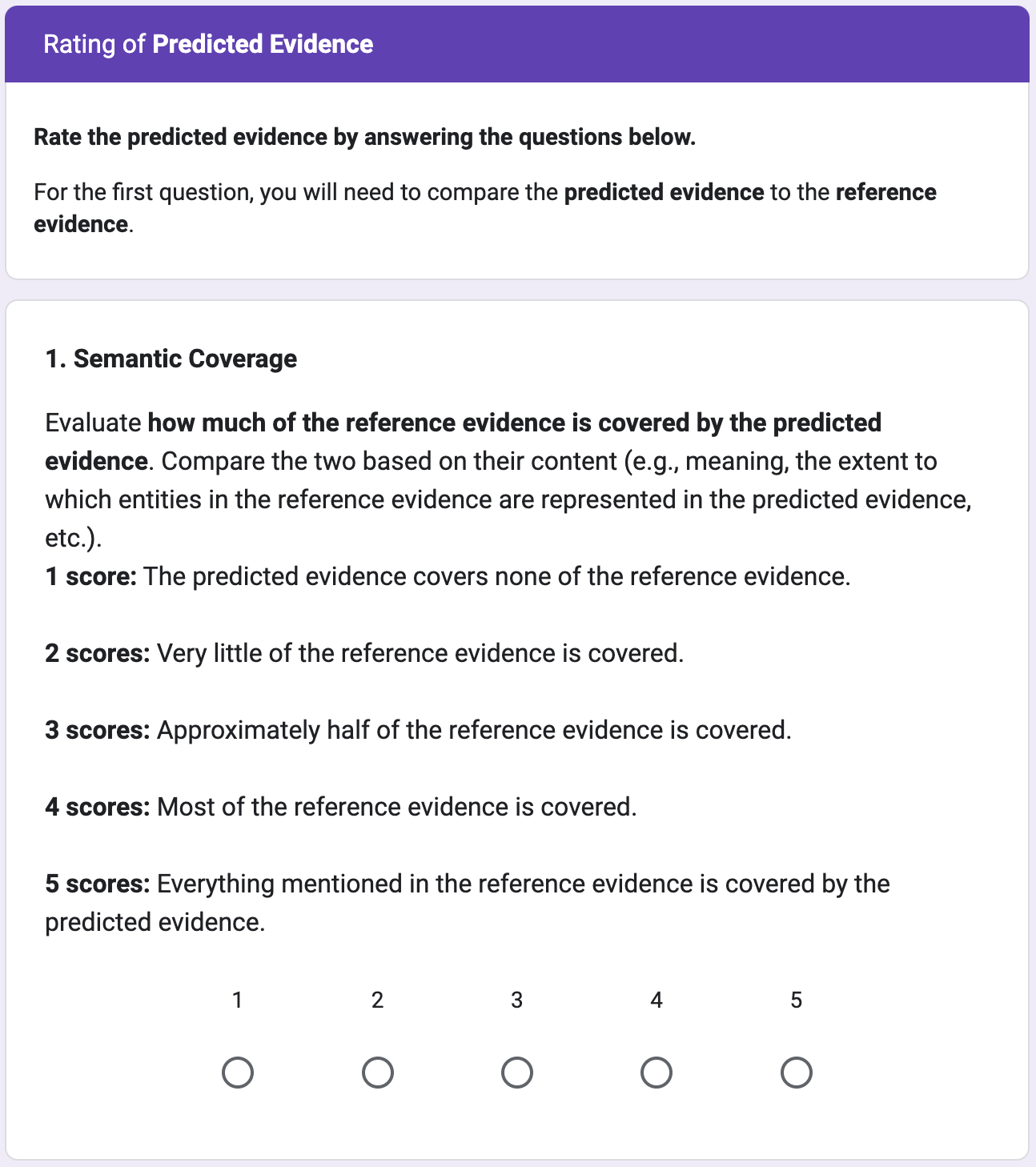}
    \caption{Platform for human evaluation of retrieved evidence from participating systems.}
    \label{fig:test4}
\end{figure*}

\begin{figure*}[ht]
    \centering
    \includegraphics[scale=0.6]{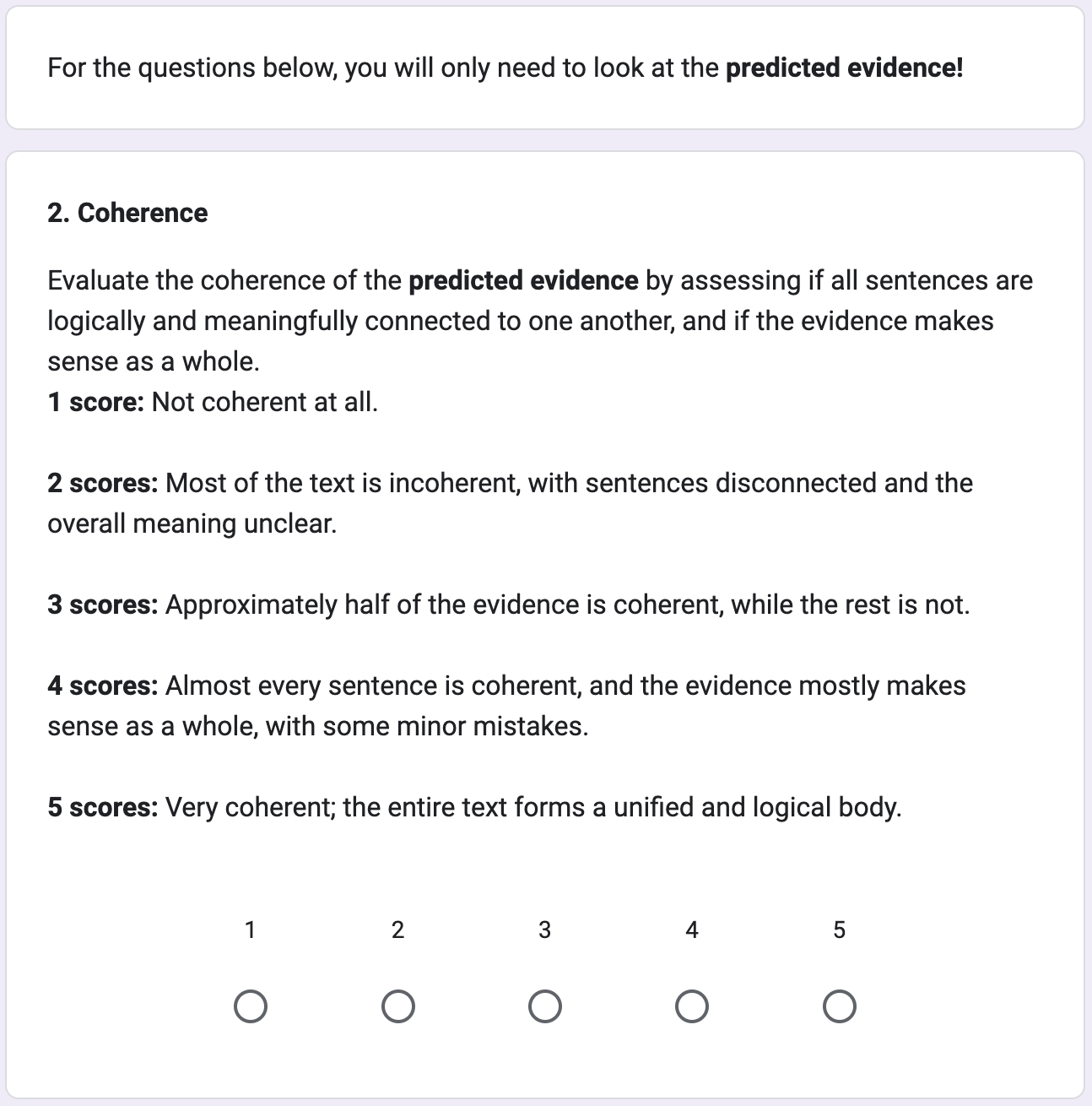}
    \caption{Platform for human evaluation of retrieved evidence from participating systems.}
    \label{fig:test5}
\end{figure*}

\begin{figure*}[ht]
    \centering
    \includegraphics[scale=0.6]{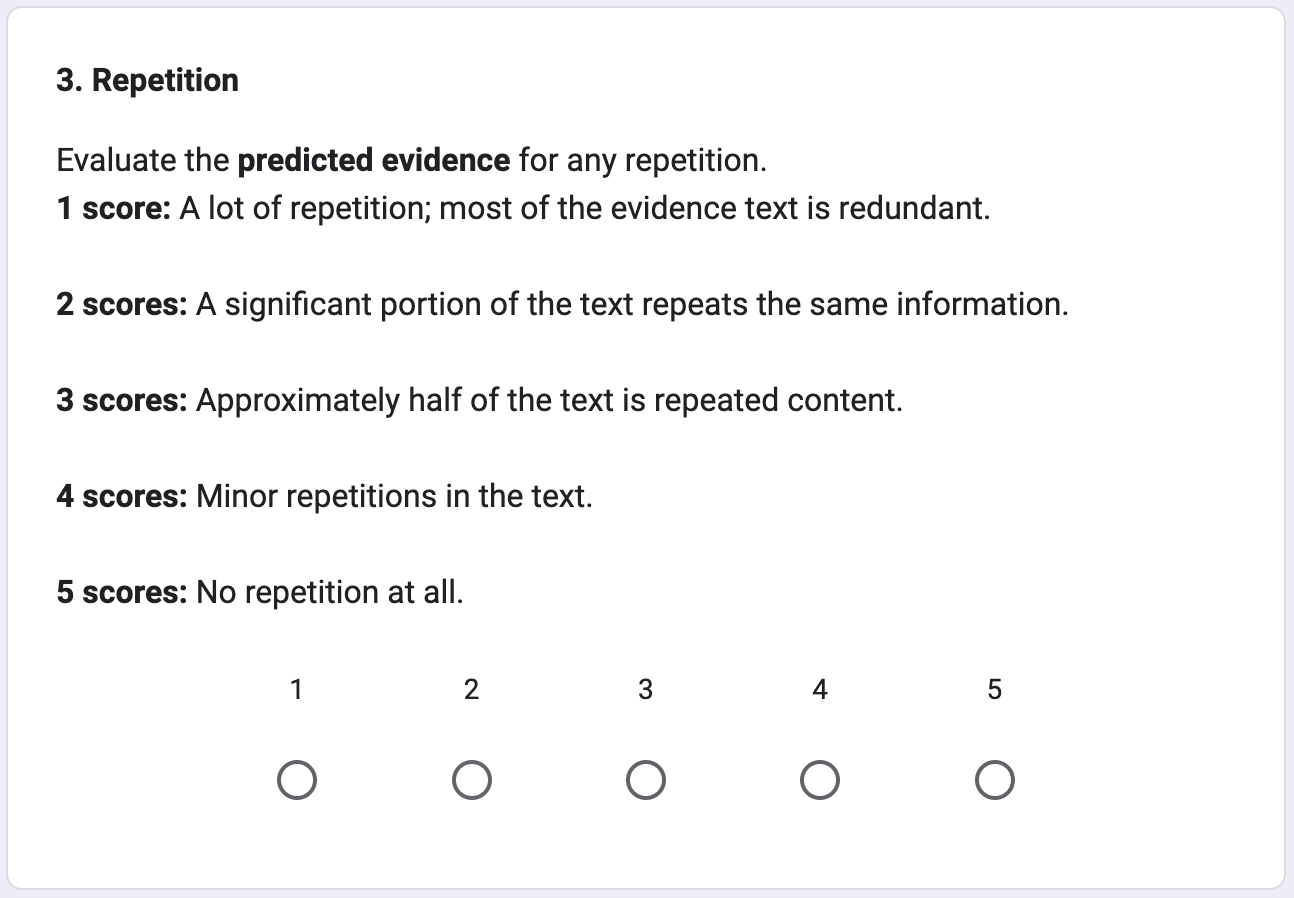}
    \caption{Platform for human evaluation of retrieved evidence from participating systems.}
    \label{fig:test6}
\end{figure*}

\begin{figure*}[ht]
    \centering
    \includegraphics[scale=0.6]{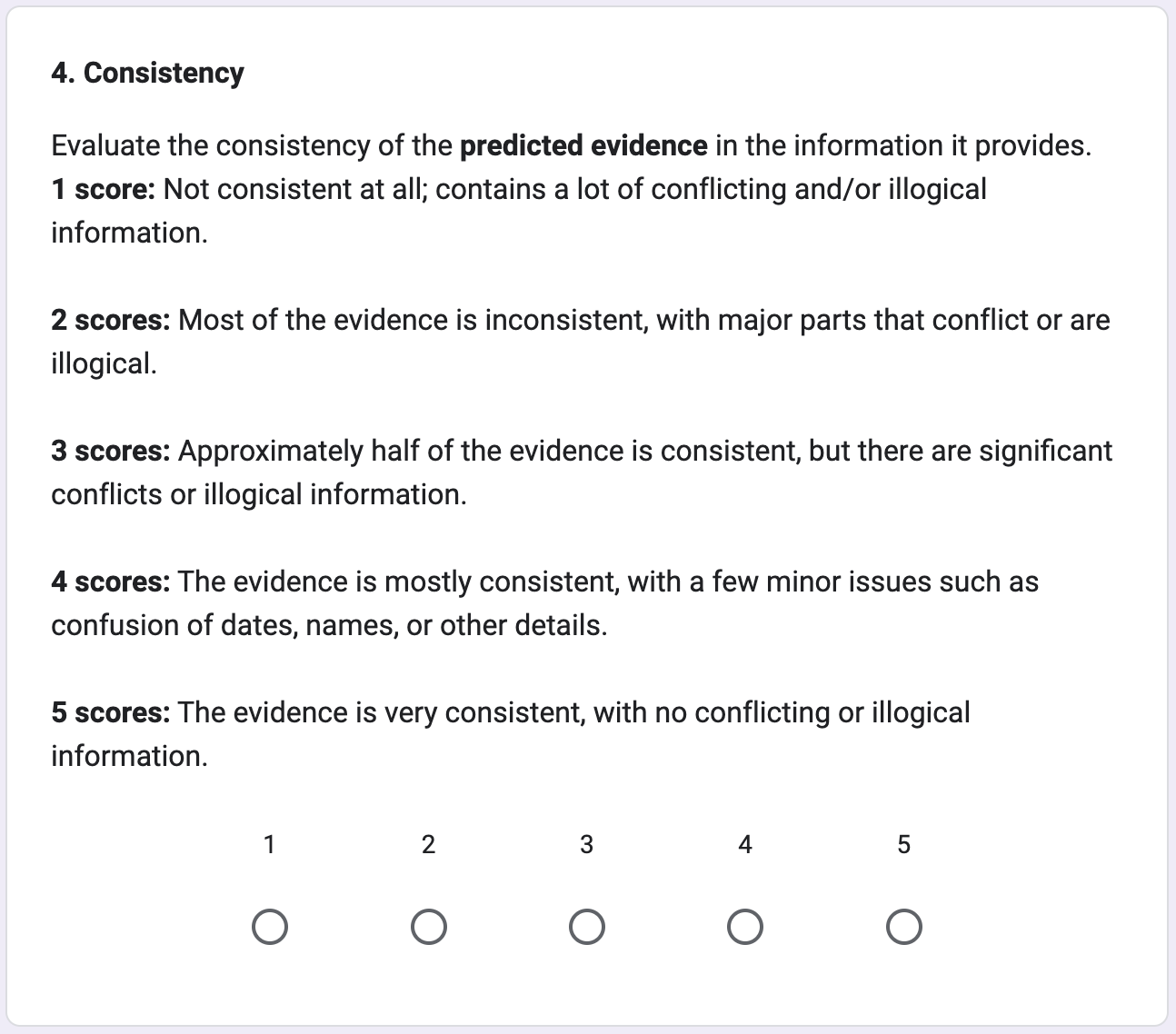}
    \caption{Platform for human evaluation of retrieved evidence from participating systems.}
    \label{fig:test7}
\end{figure*}

\begin{figure*}[ht]
    \centering
    \includegraphics[scale=0.6]{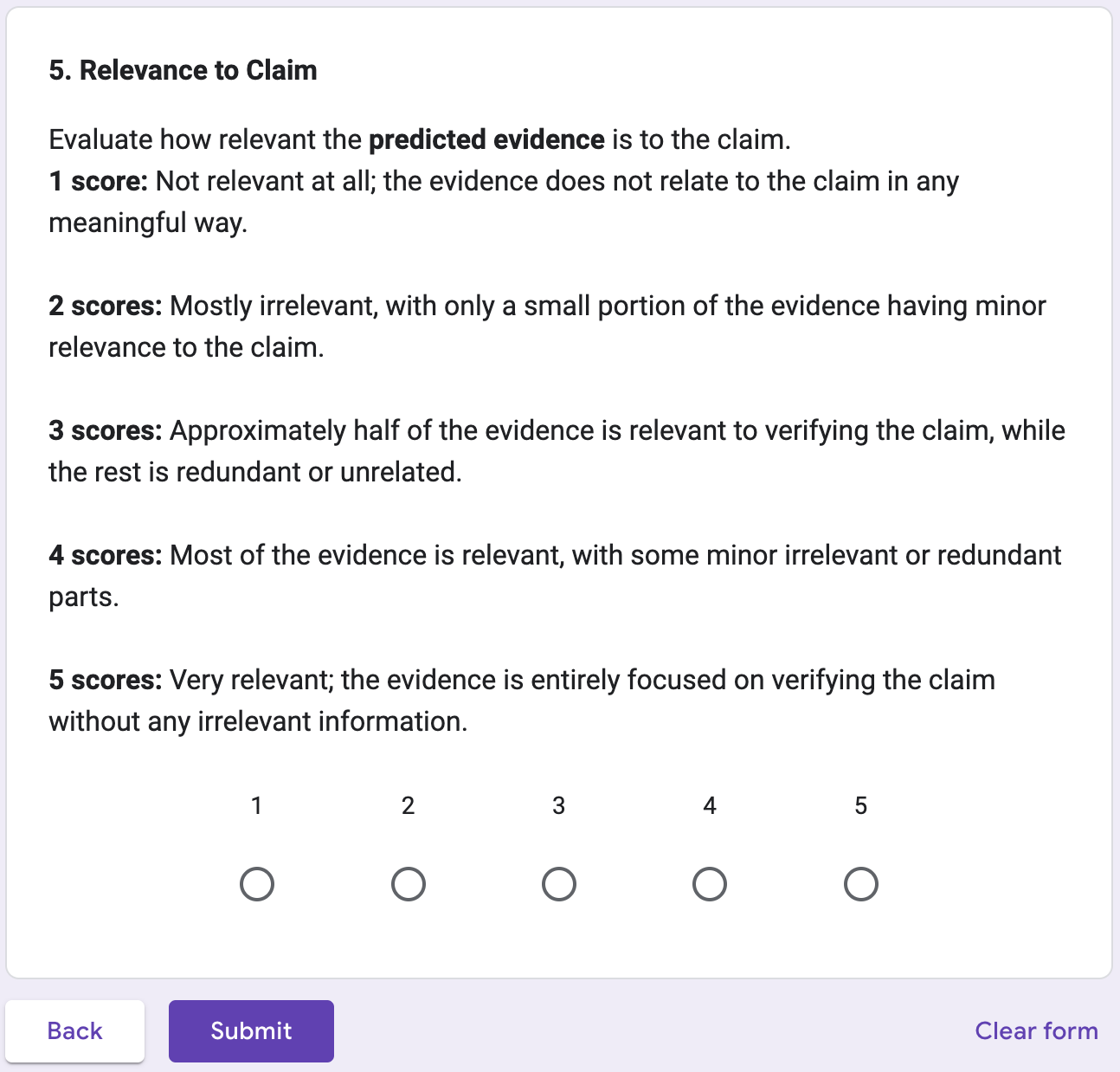}
    \caption{Platform for human evaluation of retrieved evidence from participating systems.}
    \label{fig:test8}
\end{figure*}

\begin{sidewaystable*}[t] 
    \centering
    \small
    \begin{tabular}{|p{0.3cm}|p{3cm}|p{4cm}|p{1cm}|p{1cm}|p{1cm}|p{1cm}|p{1cm}|p{1cm}|p{1cm}|p{6cm}|}
    \hline
    & \textbf{claim} & \textbf{predicted evidence} & \textbf{label human eval} & \textbf{semantic coverage} & \textbf{coher-ence} & \textbf{redun-dancy} & \textbf{consist-ency} & \textbf{rele-vance} & \textbf{pre-dicted label} & \textbf{reference evidence} \\
    \hline
    1 & New changes to the Highway Code in Great Britain states drivers should give five feet when overtaking cyclists. & Question: Should drivers give five feet when overtaking bicyclists? Answer: No answer could be found. Question: What is the Highway Code in Great Britain? Answer: No answer could be found. & not enough information & 1 & 5 & 1 & 5 & 5 & Not Enough Evidence & Question: What is 5 feet in metres? Answer: 1.5 metres. Question: What is the new overtaking distance between drivers and cyclists in the Highway Code? Answer: leave at least 1.5 metres when overtaking cyclists. Question: At what maximum speed is the new 1.5-distance rule applicable? Answer: at speeds of up to 30mph. Question: Should drivers leave more space if driving faster? Answer: Yes. On a PDF for the 'Table of Changes to the Highway Code' it states, "Give them more space when overtaking at higher speeds." \\
    \hline
    2 & Tweet portrays Ravish Kumar targeting AAP ahead of Punjab elections & Question: When were the Punjab elections? Answer: No answer could be found. Question: When was the tweet made? Answer: The tweet was allegedly made by Ravish Kumar in the midst of the Punjab election. Question: Is Ravish Kumar an AAP supporter? Answer: No answer could be found. & not enough information & 1 & 2 & 5 & 3 & 2 & Supported & Question: What does Ravish Kumar think of the AAP? Answer: No answer could be found. Question: Did Ravish Kumar tweet about the AAP? Answer: No answer could be found. Question: Has Ravish Kumar been misquoted on political matters? Answer: Senior Congress leader Digvijaya Singh apologised to NDTV anchor Ravish Kumar today after sharing a video on Twitter that incorrectly suggested the latter had abused Prime Minister Narendra Modi in a public speech on Wednesday. \\
    \hline
    3 & In 2010, a newly qualified teacher's salary in the UK was £21,500 and in 2018 it was £22,000. & Question: What was the starting salary for a newly qualified teacher in 2010? Answer: £21,500. Question: What was the starting salary for a newly qualified teacher in 2018? Answer: £22,000. Question: What was the starting salary for a newly qualified teacher in 2010 in 2018 prices? Answer: £25,400. Question: What was the starting salary for a newly qualified teacher in 2018 in 2010 prices? Answer: £23,000. & supported & 5 & 4 & 1 & 2 & 4 & Supported & Question: What was the salary of a newly qualified teacher in the UK in 2010? Answer: £21,588. Question: What was the salary of a newly qualified teacher in the UK in 2018? Answer: £23,720. Question: What was the salary of new teachers outside of London in 2010? Answer: £21,588. Question: What was salary of new teachers outside of London in 2018? Answer: Minimum starting salaries for classroom teachers will increase by between £653 (rest of England) and £816 (inner London). This means the minimum starting salary for a qualified teacher in 2019/20 will rise from £23,720 to £24,373 outside of London and from £29,664 to £30,480 in inner London. \\
    \hline
    \end{tabular}
    \caption{Examples for system predictions which are highly-ranked by the Averitec score.}
    \label{tab:evidence_evaluation_high}
\end{sidewaystable*}

\begin{sidewaystable*}[t] 
    \centering
    \small
    \begin{tabular}{|p{0.3cm}|p{3cm}|p{4cm}|p{1cm}|p{1cm}|p{1cm}|p{1cm}|p{1cm}|p{1cm}|p{1cm}|p{6cm}|}
    \hline
    & \textbf{claim} & \textbf{predicted evidence} & \textbf{label human eval} & \textbf{semantic coverage} & \textbf{coher-ence} & \textbf{redun-dancy} & \textbf{consist-ency} & \textbf{rele-vance} & \textbf{pre-dicted label} & \textbf{reference evidence} \\
    \hline
    1 & The reporting environment for foreign correspondents in China is open and free. Communication channels between foreign journalists in China and competent Chinese authorities are open and unfettered. & Question: What is the working environment for foreign journalists in China? Answer: Foreign correspondents in China have reported higher levels of concern about surveillance and invasion of privacy, as well as greater pressure by overseas Chinese officials on media organization headquarters. Nearly half of surveyed correspondents said they experienced interference, harassment, and physical violence. [...] & refuted & 5 & 5 & 5 & 5 & 5 & Refuted & Question: Does the Chinese government place restrictions on journalist content? Answer: Yes. The Propaganda Department of the Chinese Communist Party sends a detailed notice to all media every day that includes editorial guidelines and censored topics. \\
    \hline
    2 & Philippines Senator Panfilo Lacson changes his opinion and says he sees the sense in the Department of National Defense's (DND) unilateral termination of its 1989 accord with the University of the Philippines (UP). & Question: What was Philippines Senator Panfilo Lacson's previous stance on the Department of National Defense's (DND) termination of its 1989 accord with the University of the Philippines (UP)? Answer: Philippines Senator Panfilo Lacson's previous stance on the Department of National Defense's (DND) termination of its 1989 accord with the University of the Philippines (UP) was critical. [...] & supported & 3 & 5 & 5 & 5 & 4 & Supported & Question: What was his initial view on the DND and University of the Philippines longstanding arrangement? Answer: There’s sense in the move taken by the President and Sec. Lorenzana because during our red-tagging hearings in the Senate, it was established that—not only UP in particular, other universities—talagang ‘yung recruitment nanggagaling dun sa mga estudyante up to the point that they’re being killed in encounters, he told ANC's Headstart. \\
    \hline
    \end{tabular}
    \caption{Examples for system predictions which are low-ranked by the Averitec score.}
    \label{tab:evidence_evaluation_low}
\end{sidewaystable*}

\end{document}